\documentclass{article}

\usepackage{microtype}
\usepackage{graphicx}
\usepackage{subcaption}

\usepackage{amssymb}
\usepackage{microtype}

\usepackage{booktabs} 

\usepackage{hyperref}

\usepackage[accepted]{icml2023}

\usepackage{amsmath}
\usepackage{amssymb}
\usepackage{mathtools}
\usepackage{amsthm}

\usepackage{amsmath}
\usepackage{bm}
\usepackage[utf8]{inputenc} 
\usepackage[T1]{fontenc}    
\usepackage{hyperref}       
\usepackage{url}            
\usepackage{nicefrac}       
\usepackage{microtype}      
\usepackage{xcolor}         

\def\1{\bm{1}}

\def\mV{{\bm{V}}}
\def\mW{{\bm{W}}}

\DeclareMathAlphabet{\mathsfit}{\encodingdefault}{\sfdefault}{m}{sl}
\SetMathAlphabet{\mathsfit}{bold}{\encodingdefault}{\sfdefault}{bx}{n}

\usepackage{tikz}

\usepackage{hyperref}
\usepackage{url}
\usepackage{multirow}
\usepackage{enumitem}

\usepackage{amsthm}
\usepackage{amssymb}
\usepackage{mathtools}
\theoremstyle{plain}

\usepackage{varwidth}
\usepackage{enumitem}
\usepackage{tikz}
\usetikzlibrary{positioning}
\usetikzlibrary{shapes}
\usetikzlibrary{fit}
\usetikzlibrary{calc}

\usepackage{listings}
\usepackage{xcolor}
\usepackage{courier}

\definecolor{codegreen}{rgb}{0,0.6,0}
\definecolor{codegray}{rgb}{0.5,0.5,0.5}
\definecolor{codeblack}{rgb}{0.,0.,0.}
\definecolor{codepurple}{rgb}{0.58,0,0.82}
\definecolor{backcolour}{rgb}{0.95,0.95,0.92}

\lstdefinestyle{mystyle}{
    backgroundcolor=\color{backcolour},   
    commentstyle=\color{codegreen},
    keywordstyle=\color{codeblack},
    numberstyle=\tiny\color{codegray},
    stringstyle=\color{codepurple},
    basicstyle=\ttfamily\footnotesize,
    breakatwhitespace=false,         
    breaklines=true,                 
    captionpos=b,                    
    keepspaces=true,                 
    numbers=left,                    
    numbersep=5pt,                  
    showspaces=false,                
    showstringspaces=false,
    showtabs=false,                  
    tabsize=2,
    aboveskip=0pt,
    belowskip=-3pt
}

\usepackage{pifont}
\usepackage[capitalize,noabbrev]{cleveref}

\newenvironment{itemize*}%
  {
  \setlength{\parskip}{0pt}
  \begin{itemize}%
    \setlength{\itemsep}{5pt}
    \setlength{\parskip}{0pt}
    \setlength{\parsep}{0pt}
    }%
  {\end{itemize}\setlength{\parskip}{11pt}
}

\theoremstyle{plain}
\theoremstyle{definition}

\theoremstyle{remark}

\usepackage[textsize=tiny]{todonotes}

\icmltitlerunning{Expand or Narrow your representation}

\begin{document}

\twocolumn[
\icmltitle{A surprisingly simple technique to control the pretraining bias for better transfer: Expand or Narrow your representation}









\begin{icmlauthorlist}
\icmlauthor{Florian Bordes}{comp,mila}
\icmlauthor{Samuel Lavoie}{comp,mila}
\icmlauthor{Randall Balestriero}{comp}
\icmlauthor{Nicolas Ballas}{comp}
\icmlauthor{Pascal Vincent}{comp,mila}
\end{icmlauthorlist}

\icmlaffiliation{comp}{Meta AI}
\icmlaffiliation{mila}{Mila, Universite de Montreal}

\icmlcorrespondingauthor{Florian Bordes}{florian.bordes@umontreal.ca}

\icmlkeywords{Machine Learning, ICML}

\vskip 0.3in
]



\printAffiliationsAndNotice{}  

\begin{abstract}
Self-Supervised Learning (SSL) models rely on a pretext task to learn representations. Because this pretext task differs from the downstream tasks used to evaluate the performance of these models, there is an inherent misalignment or \textit{pretraining bias}. A commonly used trick in SSL, shown to make deep networks more robust to such bias, is the addition of a small projector (usually a 2 or 3 layer multi-layer perceptron) on top of a backbone network during training. In contrast to previous work that studied the impact of the projector architecture, we here focus on a simpler, yet overlooked lever to control the information in the backbone representation. 
We show that merely changing its dimensionality -- by changing only the size of the backbone's very last block -- is a remarkably effective technique to mitigate the pretraining bias. It significantly  
improves downstream transfer performance for both Self-Supervised and Supervised pretrained models.
\end{abstract}

\section{Introduction}
\label{sec:intro}

\begin{figure*}[ht]
    \centering
    \includegraphics[scale=0.55]{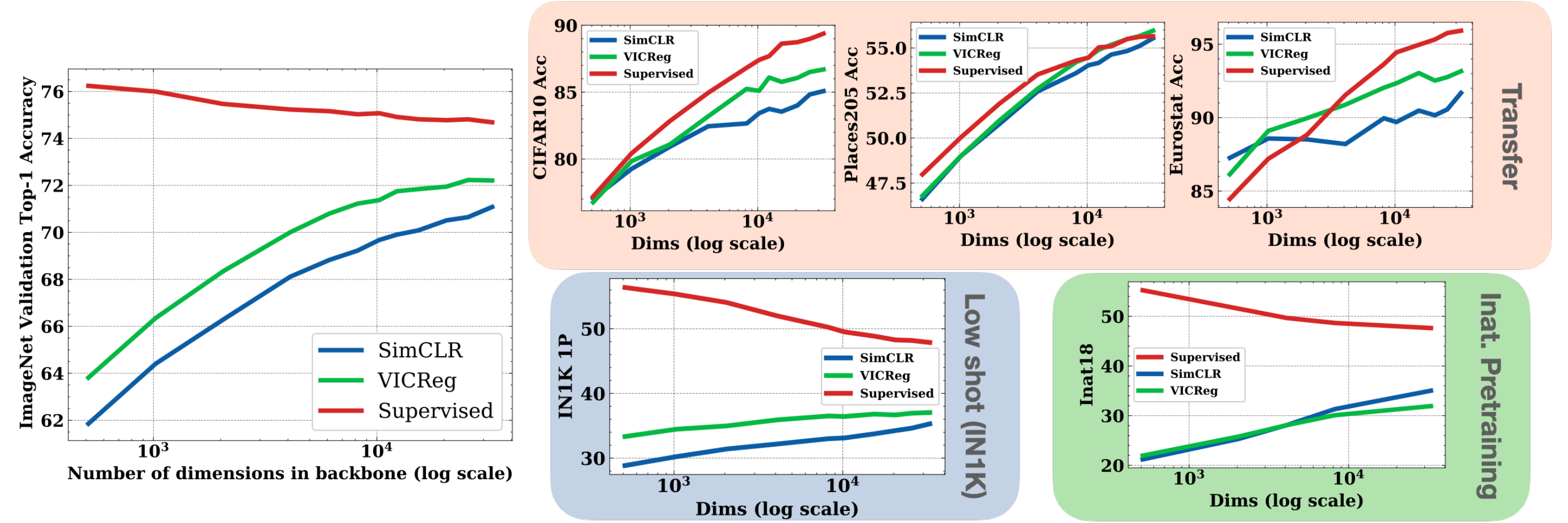}
    \caption{We study how changing the backbone dimension $D$ (by changing the number of features in the last convolutional block of a Resnet50) impacts  performance in Supervised and Self-Supervised learning. All the models were pretrained using ImageNet, except the models presented in the bottom right green plot in which pretraining was performed using the Inat18 dataset. All the evaluations were done with linear probing. In the leftmost plot we show the performances on the validation set of ImageNet. SSL methods (SimCLR and VICReg) achieve better performance when increasing $D$. This is not the case for supervised pretrained model, that even sees a drop of 10\% accuracy in low-shot evaluation. By contrast, when looking at the performances in transfer, we observe that all methods, whether supervised or self-supervised, achieve better performances when increasing $D$. When the pretraining and downstream tasks are perfectly aligned (training and evaluating on ImageNet on the same classification task), then the model will benefit from learning strong invariances by using a smaller $D$, however when there is a misalignement, too strong invariance can hurt the model performance since important information that is needed to solve the downstream task might have been removed.}
\end{figure*}

The self-supervised learning (SSL) paradigm aims at learning representations by using "pretext tasks", such as solving Jigsaw-puzzles \citep{Noroozi_2016_Jigsaw}, predicting rotations \citep{Gidaris2018arxiv}, denoising or recovering partially-masked input~\citep{denoising_vincent, he2022masked}, or encouraging learned representations to be invariant to a set of handcrafted representations \citep{chen2020simclr, caron2020swav, zbontar2021barlow}. SSL methods will tend to more or less \textit{overfit} their pretext task, which induces a \emph{pretraining bias} with respect to the downstream (transfer) task of interest -- and the representation that would be best for it \citep{Guillotine}.
Since such pretraining bias hurts downstream task performance, a common trick (named \textit{Guillotine Regularization} in \citet{Guillotine}) is to add a projector on top of a backbone network during training and discard it during evaluation.
This trick can mitigate the detrimental effect of the pretraining bias only to some degree. Despite having shown great success on relatively class-balanced datasets such as ImageNet-1k \citep{deng2009imagenet}, ~\citet{tian2021divide} demonstrated that SSL methods trained on long-tailed datasets perform poorly, in contrast with supervised baselines. ~\citet{uniform_prior} argued that such failure of SSL methods is attributable to a specific bias: an implicit uniform prior in their objective. To correct for this, the authors show how to modify their clustering-based SSL method, called MSN~\citep{msn}, to change cluster priors to more closely follow the class distribution. Unfortunately, many SSL methods are not based on an explicit clustering, hence the proposed change is not applicable to them. 

In this paper we study a simple generic way to further improve robustness with respect to task misalignment and the pretraining bias occurring in SSL -- including the bias due to an implicit uniform prior. We will show that merely expanding or narrowing the backbone dimension -- a currently unexploited lever\footnote{Largely unexplored and unexploited in SSL because backbones  architectures are typically extracted as is from supervised models, and left untouched.} -- is remarkably effective at controlling the pretraining bias. It allows to significantly improve transfer performance, as well as robustness when training on datasets with long-tailed class distribution. Our main contributions highlight that:


\begin{itemize*}
    \item Training SimCLR with a very small linear projector (32 neurons) can lead to competitive results on ImageNet. 
    \item SSL methods strongly benefit from significantly wider backbone representations. This calls into question the widespread use in SSL of fixed backbone architectures that were designed to work best for supervised tasks.
    \item By contrast, supervised training yields better in-distribution test accuracy when using smaller representations. However when pretraining a supervised model destined to transfer tasks, it is better to use a larger backbone representation, to mitigate the pretraining bias, similarly to SSL.
    \item Wider SSL representations are extremely sparse. It's possible to binarize them without any significant loss in performance. 
\end{itemize*}

\section{Related work}

\paragraph{Analysis of the representation's dimensionality in SSL} Several works have studied the impact of the projector architecture in SSL. Some methods like SimCLR \citep{chen2020simclr} achieve similar performance when using small or large projector embeddings, while methods like VICReg are sensitive to the projector's dimensionality \citep{garrido_duality_2022}. In this paper, we primarily focus on the effect of the backbone dimensionality. A closer and more recent work ~\citet{dubois_improving_2022} observed that larger backbone representations lead to better linear probe performance when using CISSL. We generalize this result to SimCLR, VICReg, Byol and the supervised setting and shed light on the importance of wider representations to mitigate the pretraining bias. In addition, we provide a deeper quantitative and qualitative analysis of the impact of larger representations with respect to several downstream tasks and pretraining configurations. 

\paragraph{The hidden uniform prior} \citet{uniform_prior} demonstrate the existence of a \textit{hidden} uniform prior when training contrastive self-supervised model. By linking many SSL methods to K-means clustering, they shed light on how this prior enables the uniform clustering of the data in the representation space. They show that such prior can be harmful when using imbalanced datasets. 
In the present paper, we argue that this implicit uniform prior is most problematic due to the dimensional bottleneck being usually too strong, thus inducing the selection of features that are not aligned with the downstream tasks.

\paragraph{Sparsity and Binarization in neural networks} Several works and methods have been proposed to encourage the sparsity of 
representations through specific training criteria \citep{sparse_coding, yann_sparsity}. 
By contrast, our paper shows that high degrees of sparsity (> 80\%) can emerge naturally in SSL-trained, post-ReLU representations, without any additional constraint or criterion, when using wider representations. In addition, we show that sparse activations in SSL make the representation easily binarizable, without any significant performance loss.

\begin{figure}[ht]
    \centering
     \begin{subfigure}{0.25\textwidth}
         \centering
         \includegraphics[scale=0.26]{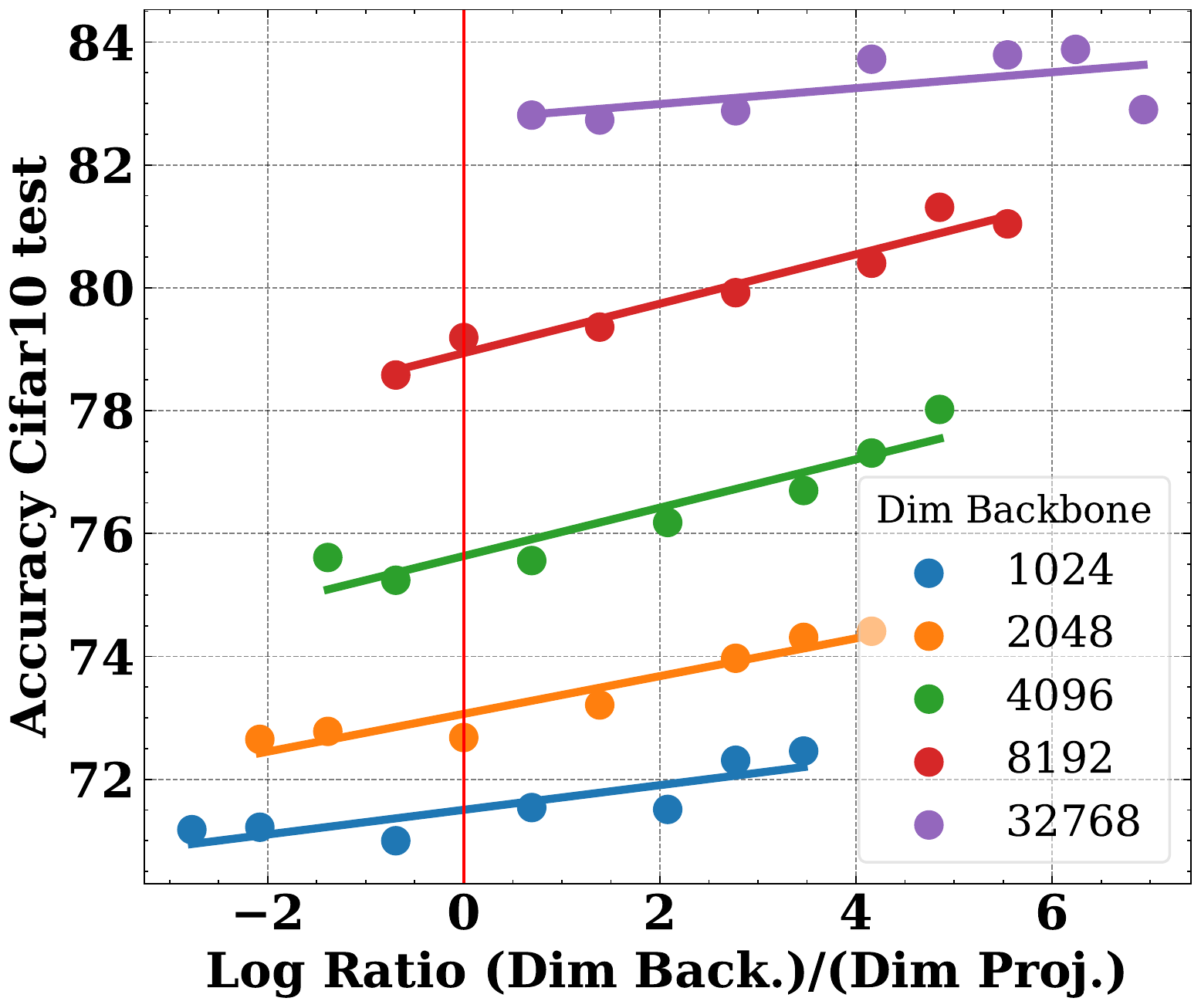}
         \caption{Unbalanced sampling}
         \label{fig:linear_cifar10_unbalanced}
     \end{subfigure}%
    \begin{subfigure}{0.25\textwidth}
         \centering
         \includegraphics[scale=0.26]{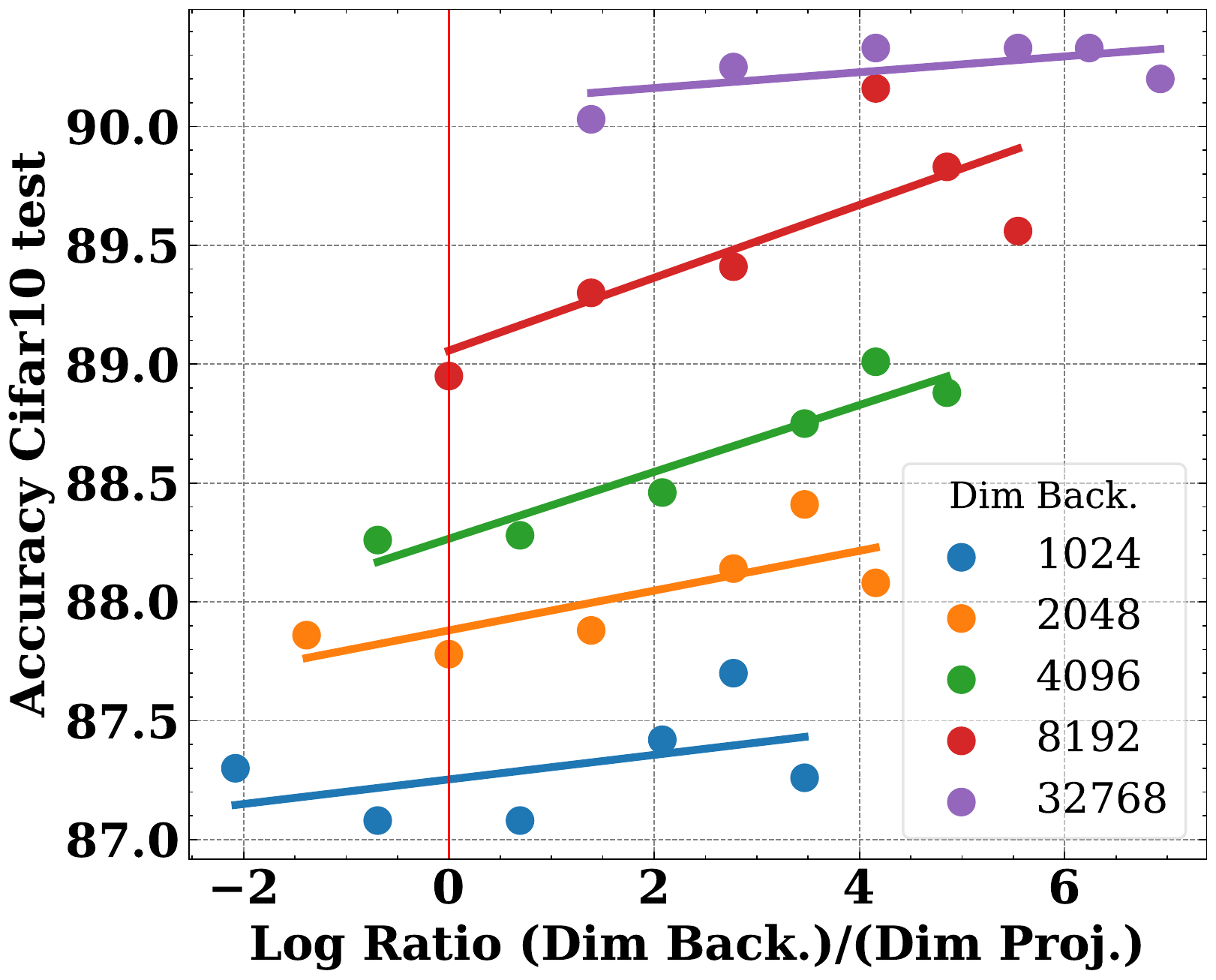}
         \caption{Balanced sampling}
         \label{fig:linear_cifar10_balanced}
     \end{subfigure}
    \caption{Accuracy on the test set of CIFAR10 with SimCLR trained with a \textbf{linear projector} on 300 epochs for different backbone dimension $D$ (from 1024 to 32768) and different linear projector dimension $K$ (from 32 to 16362). When $D>K$ (positive log ratio), the performance is evidently improved for any given $D$. In both unbalanced and balanced cases, \textbf{increasing the backbone dimension for any given projector dimension leads to significant improvement in terms of accuracy.}} 
    \label{fig:linear_cifar10}
\end{figure} 
    
\begin{figure}[ht]
    \centering
     \begin{subfigure}{0.25\textwidth}
         \centering
         \includegraphics[scale=0.26]{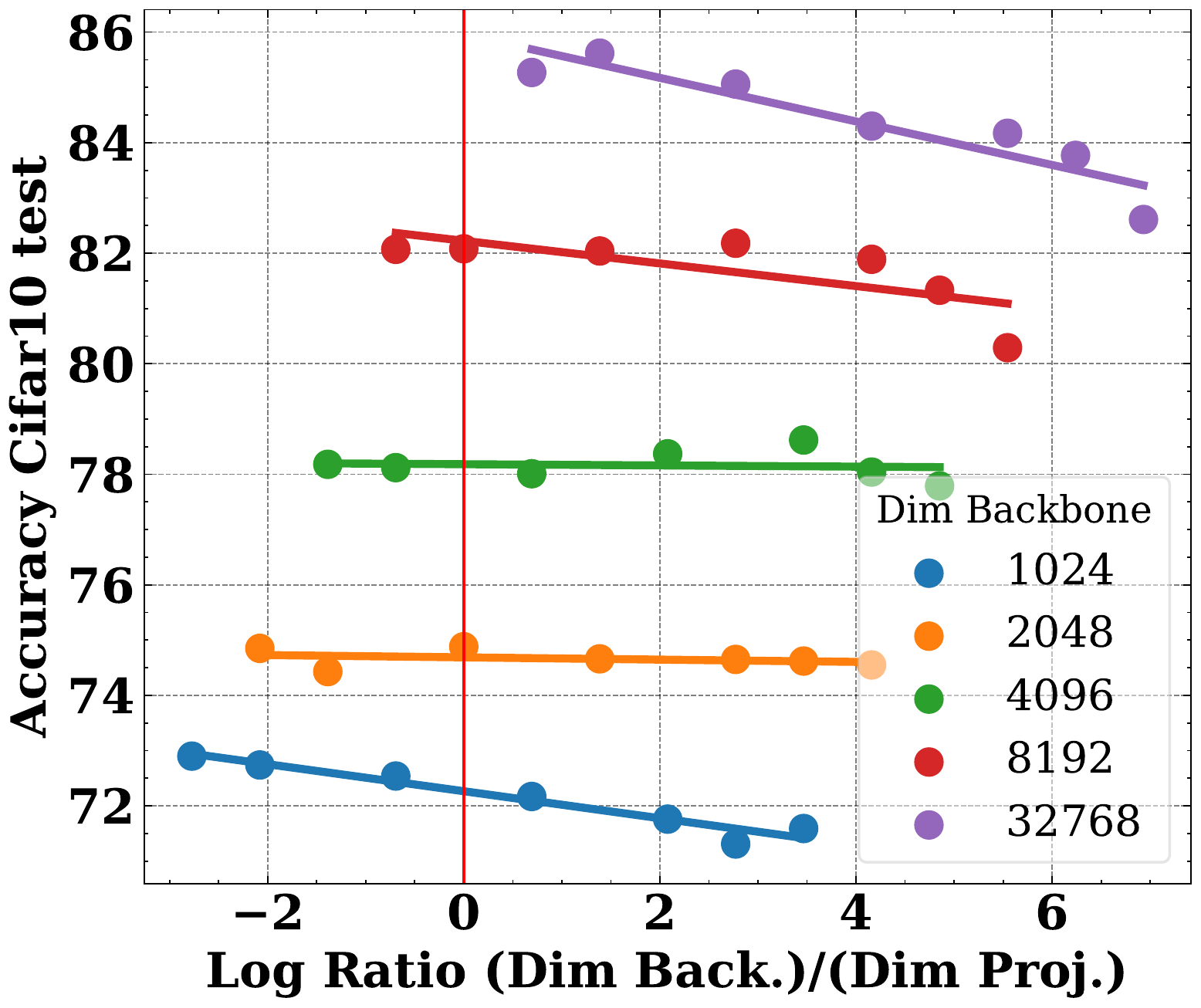}
         \caption{Unbalanced sampling}
         \label{fig:nonlinear_cifar10_unbalanced}
     \end{subfigure}%
    \begin{subfigure}{0.25\textwidth}
         \centering
         \includegraphics[scale=0.26]{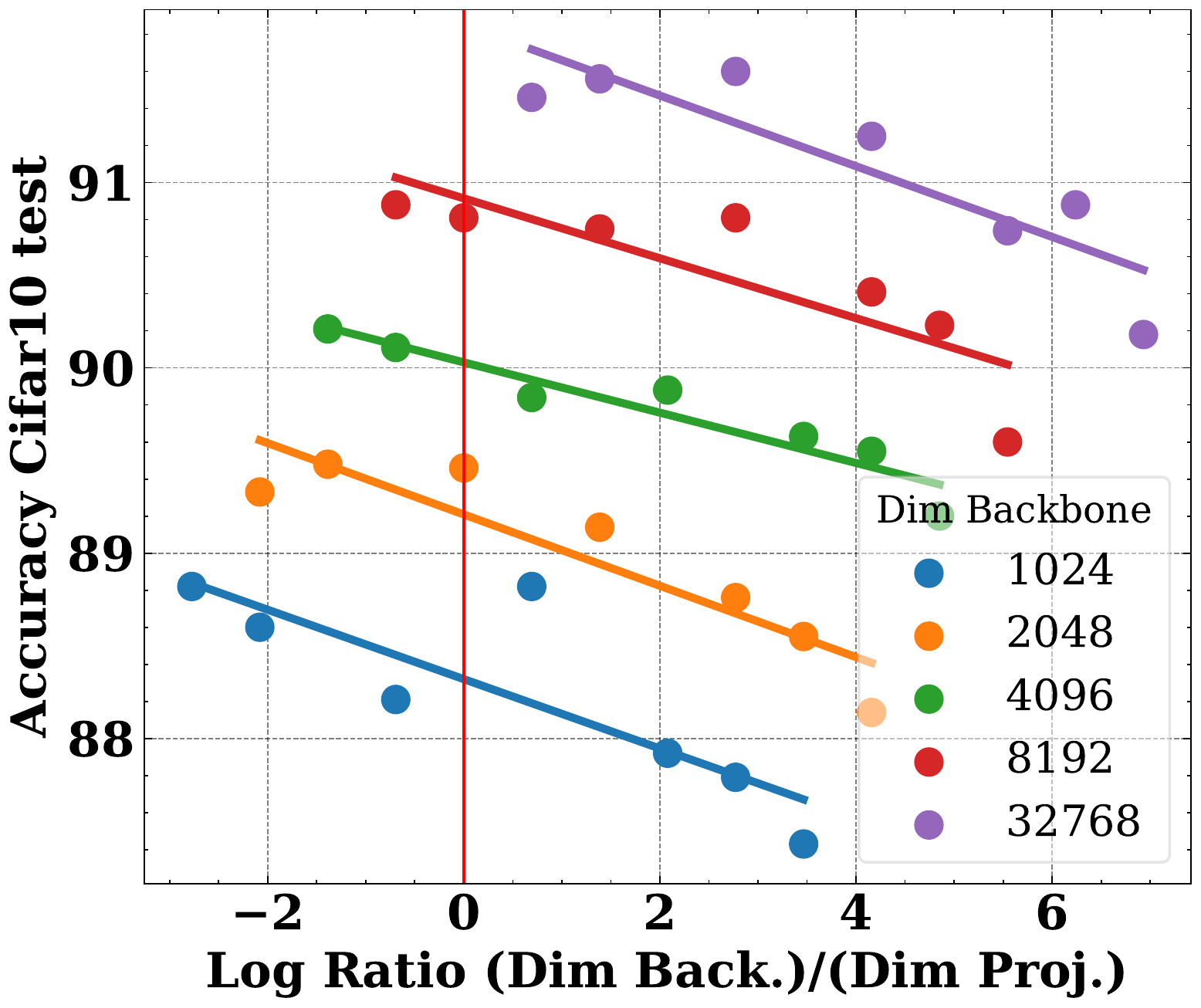}
         \caption{Balanced sampling}
         \label{fig:nonlinear_cifar10_balanced}
     \end{subfigure}

    \caption{Accuracy on the test set of CIFAR10 with SimCLR trained with a \textbf{nonlinear projector} on 300 epochs for different backbone dimension $D$ (from 1024 to 32768) and different first linear projector layer dimension $K$ (from 32 to 16362) while keeping the second projector dimension fixed to 256. In contrast to \cref{fig:linear_cifar10}, for a given size of backbone, increasing the ratio $D/K$ does not improve the performance. However, similarly to \cref{fig:linear_cifar10} in both unbalanced and balanced cases, \textbf{increasing the backbone dimension for any given number of projector dimension leads to significant accuracy improvements.}} 
    \label{fig:control_cifar10}
\end{figure}

\section{Controlling the information bottleneck}

We consider a deep network \emph{backbone} as a function $f:\mathcal{I} \mapsto \mathbb{R}^{D}$, that maps an image $x \in \mathcal I$ to a $D$-dimensional \emph{backbone representation}. In the considered SSL methods, this representation is further transformed by a \emph{projector} $g:\mathbb{R}^{D} \mapsto \mathbb{R}^{K}$ into a $K$-dimensional \emph{projector embedding}. In joint-embedding SSL methods, the loss $\mathcal{L}$ depends on a minibatch of images and their augmentations, jointly denoted $X$. We thus optimize $\mathcal{L}(g(f(X)))$.

\subsection{Using a linear projector}

To get insights on how to leverage the projector and backbone's dimensionality to improve robustness to the inherent misalignment between the SSL objectives and the data, we start by studying a very simple case with SimCLR by using a linear projector. Let $\mW \in \mathbb{R}^{D \times K}$ be the linear projector matrix (for simplicity we omit the bias as it does not change anything to our analysis). The optimized loss thus becomes
$\mathcal{L}(\mW f(X))$, and
in this setting the gradient signal backpropagagted to the backbone $f$ is given by
\begin{align*}
    \left[\nabla \mathcal{L}(\mW f(X))\right]\mW^T,
\end{align*}
We see from this equation that the backpropagated signal is constrained to live in a subspace of dimension at most $\mathrm{rank}(\mW)=\min(D, K)$. To limit the amount of ``gradient-based information'' coming from the loss, one must have $D\gg K$ which creates 
 an information bottleneck. To see that, imagine the simple scenario of $f$ also being linear and initialized full-rank i.e. the input-output mapping is
\begin{align*}
    \mW\mV X,
\end{align*}
where we omit bias terms for clarity and where $\mV \in \mathbb{R}^{d \times D}$ i.e. the input space is of dimension $d$ and is commonly $d \gg D$. Therefore, when $D>K$ we can express the above as
\begin{align*}
    \mW(\mV_1+\mV_2) X,
\end{align*}
where $\mV_1$ is constrained to live in the $K$ dimensional subspace spanned by $\mW^T$ and $\mV_2$ in its orthogonal space. From that, we directly see that a gradient update, only living within $\mV_1$'s space, will leave $\mV_2$ at its initial value. Let us now consider what happens if projector $\mW$ is frozen, and we train $\mV$ from a random initialization. We see that the smaller $K$ is, the larger the subspace of $\mV$ that will be left untouched at its initial value; so that $\mV X$ will retain the most information about $X$ (as a mostly random projection), barely affected by the SSL loss $\mathcal{L}$.  


The above discussion was meant to build insight on the behavior in the simplest setting. We now want to verify whether this insight translates experimentally to a realistic model. For our experiments with linear projectors, we use a regular nonlinear Resnet50 \citep{he2016resnet} backbone. To effectively change the dimension $D$ of the backbone, we add a hyperparameter that can increase by a given factor $\alpha$ the number of feature maps in its last convolutional block. We choose to increase the size of the bottleneck \emph{before pooling} in order for the network to be able to retain more spatial information about the input image. This method can be applied on both supervised and self-supervised architectures, since it is performed on the backbone common to both. When changing the dimension of the linear projector, we only change the number of output units of the linear layer.

\begin{table}[h]
    \centering
    \footnotesize
    \caption{\textbf{Changing the linear projector dimension for a fixed backbone dimension (32768) on ImageNet and Inat} with SimCLR. The best performances are obtained with a very small linear projector of size 32. The network used to compute the performances on ImageNet was pretrained using ImageNet while the network used to compute the performances on Inat18 was pretrained on Inat18.
    }
    \label{tb:imagenet_linear}
    \begin{tabular}{r||c|c|c|c}
        \hline
        Nb. of Dims. & 0 & 32 & 8192 & 16384\\
        \hline
        ImageNet Val. Acc & 46.7 & \textbf{67.4} & 65.5 & 65.3 \\
        \hline
        Inat Val. Acc & 4.1 & \textbf{27.76} & 23.9 & 23.9 \\
        \hline
    \end{tabular}
\end{table}

\paragraph{Linear projector with unbalanced sampling} To experimentally assess how much the ratio $D/K$ increases the robustness with respect to the uniform prior, we design a similar toy experiment as in \citet{uniform_prior} in which the sampling distribution changes during training. Instead of sampling uniformly from the CIFAR10 dataset, we sample images belonging to only two different classes per mini batch. Then, we use a Resnet50 with a SimCLR criterion in which we vary the backbone dimension $D$ and projector dimension $K$. After SSL pre-training, we train a downstream linear probe on the 
backbone's output, using the traditional uniform sampling strategy. We present the results in \cref{fig:linear_cifar10_unbalanced} which shows improvement in accuracy when $D\gg K$. For a given backbone dimension $D$, decreasing the projector dimension $K$ improves the performances by a few percentage points. For a given projector dimension $K$, increasing the backbone dimension leads to even higher gains. This figure shows that the best performances are obtain for a very large $D$ and small $K$. 

\paragraph{Linear projector with balanced sampling} Since changing the ratio and the backbone dimension yields significant improvements in the unbalanced data experiment, we now turn to the balanced scenario in which we sample uniformly from the CIFAR10 dataset in \cref{fig:linear_cifar10_balanced}. In this setting there appears to be no benefit in increasing or decreasing the number of dimension in the projector for a given backbone size. However, there is still a significant gain in accuracy when increasing the size of the backbone for a fixed projector dimension. To explore if this last observation holds on a larger scale setup, we used ImageNet \citep{deng2009imagenet} in which we fixed the backbone dimension to 32768 and change only the projector dimension. The results are available in \cref{tb:imagenet_linear} and show that decreasing the size of the linear projector allows us to improve the validation accuracy on ImageNet. It is worth noting that it is possible to get a \textbf{SimCLR model that reaches $67.4\%$ accuracy on ImageNet with a single linear projector using only 32 dimensions.}

\subsection{Using a nonlinear projector}

\paragraph{Nonlinear projector with unbalanced sampling} Since the linear projector experiment shows that increasing the backbone dimension leads to significant performance improvements with SimCLR when trained on unbalanced data, we extend the previous experiment to the nonlinear case. We start by using again our setup on CIFAR10, except that we use a two-layer nonlinear projector, where the first layer is a mapping to $K$ dimensions, while the second layer is fixed to 256 dimensions (the value used in the original work of \citet{chen2020simclr}). In \cref{fig:nonlinear_cifar10_unbalanced}, we show the performances with this nonlinear projector for various ratios $D$/$K$. In contrast with the experiments using the linear projector there doesn't seem to be any benefit in changing the ratio $D/K$ for a fixed $D$. However, increasing the value of $D$ for a fixed $K$ still leads to significant gains in terms of accuracy. 

\begin{table}[ht]
    \centering
    \footnotesize
    \caption{\textbf{Imbalanced data on ImageNet:} In this experiment, we train a Supervised model, SimCLR and VICReg with various backbone dimensions using imbalanced mini batches following a similar setup as \citet{uniform_prior}. Instead of filling the mini batches during training by sampling uniformly from the dataset, we change the sampling distribution such that we get only 8 different classes per mini batches. Then, for a given backbone dimension, we measure the difference in accuracy between a model trained with class balanced and class imbalanced mini batches. We observe for both SSL models that their performances dramatically decrease when having narrower backbone dimension. Using larger backbone dimension significantly reduces the gap.
    \vspace{.2cm}
    }
    \label{tb:class_stratified_sampling_imagenet}
    \begin{tabular}{r|ccccc|cc}
        \toprule
        & 512 & 2048 & 8192 & 32768\\
        \toprule
        Supervised & \color{red}-6.54 & \color{red}-2.65 & \color{red}-0.84 & \color{red}-0.5 \\
        SimCLR & \color{red} -14.09 & \color{red}-11.0 & \color{red}-9.2 & \color{red}-7.5 \\
        VICReg & \color{red}-17.28 & \color{red}-14.5 & \color{red}-11.9 & \color{red}-9.8 \\
         \bottomrule
    \end{tabular}
\end{table}

\begin{table}[ht]
    \centering
    \footnotesize
    \caption{\textbf{Imbalanced data on Inat:} In this experiment, we train a Supervised model, SimCLR and VICReg on the dataset Inaturalist18 \citep{inat18} containing 8,142 classes. We observe for both SSL models that their performances dramatically decrease when having narrower backbone dimension. Using larger backbone dimension reduces significantly the gap.
    \vspace{.2cm}
    }
    \label{tb:class_stratified_sampling_inat}
    \begin{tabular}{r|ccccc|cc}
        \hline
        & 512 & 2048 & 4096 & 8192 & 32768\\
        \hline
        Supervised & \textbf{55.25} & 51.53 & 49.66 & 48.66 & 47.64 \\
        VICReg & 21.95 & 25.8 & 28.09 & 30.09 & \textbf{31.92} \\
        SimCLR & 21.20 & 25.33 & 28.10 & 31.32 & \textbf{35.03} \\ 
         \hline
    \end{tabular}
\end{table}

\begin{figure}[ht]
    \centering
    \includegraphics[scale=0.38]{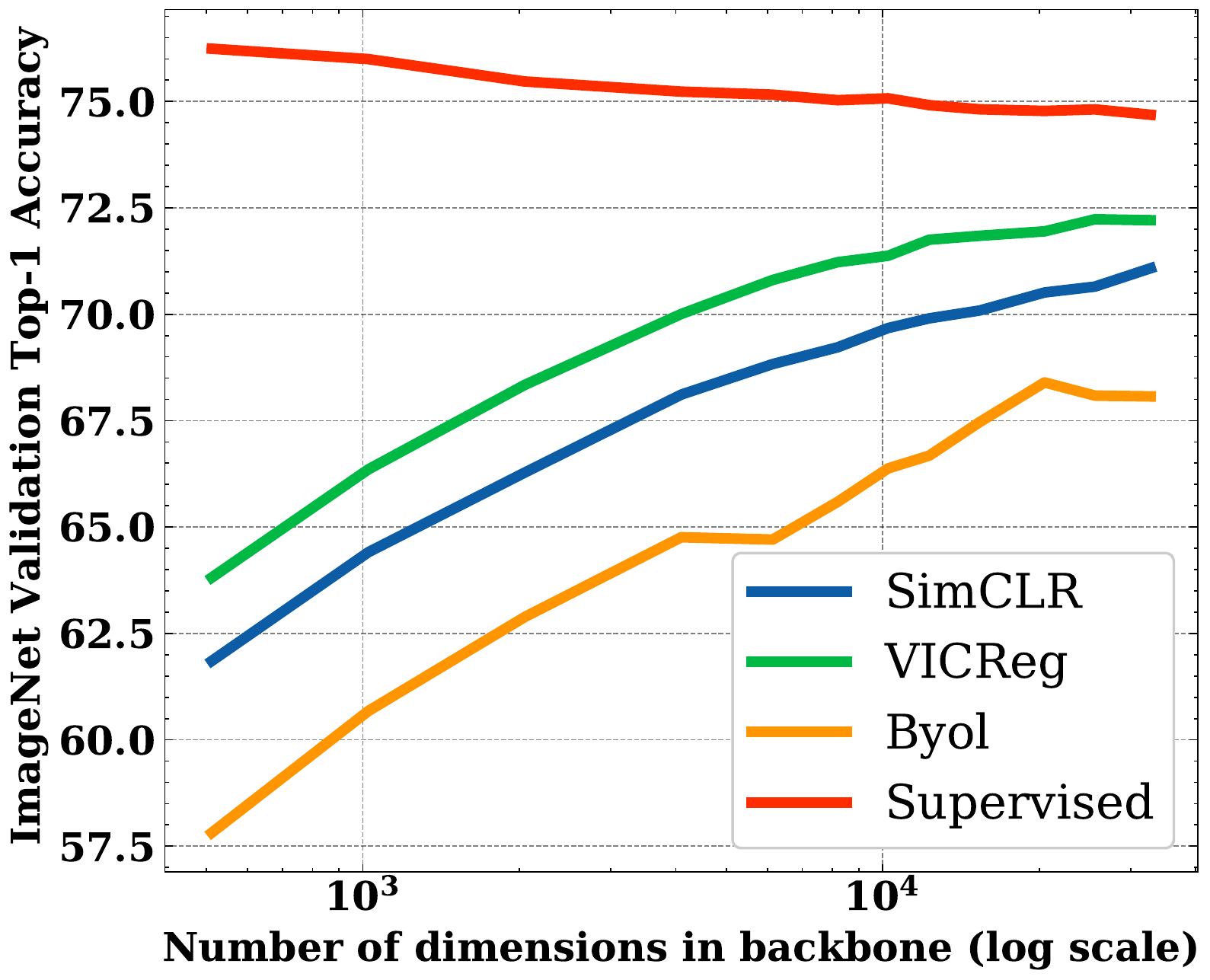}
    \caption{Validation accuracy on ImageNet\citep{deng2009imagenet} with linear probing for self-supervised and supervised models trained using different representation sizes for 100 epochs. We modify the last convolutional block of a Resnet50 to output a specific number of feature maps that we vary from 512 to 32768. Using this Resnet, we train the following self-supervised models: SimCLR, VICReg and Byol. In this figure, we observe that the traditional supervised baseline don't benefit at all from increased backbone size (but actually benefit when decreasing it !) whereas most SSL models using a projector actually benefit from wider representation. } 
    \label{fig:diff_acc_bin_cont}
\end{figure}

\begin{table}[ht]
    \footnotesize
    \centering
    \caption{\textbf{Transfer learning} Linear probe accuracy for SimCLR, VicReg, Byol and a supervised model across several downstream tasks. Even if the supervised model get worse performances on ImageNet when increasing the size of the backbone, it's worth to note that when looking at other downstream tasks the performances increase significantly when using wider backbone. This showed that even in the supervised learning case, increasing the dimension of the backbone can increase significantly the robustness with respect to the pretraining bias. We present an extended table with more datasets and dimensions in \cref{tab:downstream_tasks_all}. 
    \vspace{.2cm}
    }
    \label{tab:downstream_tasks}
    \resizebox{\columnwidth}{!}{%
    \begin{tabular}{|r|c|c|c|c|c|c|c|c|c|c|c}
        \hline
         \textbf{Dataset} & \textbf{Model} & \multicolumn{6}{c|}{\textbf{Backbone Dimension}}\\
          &  & 512 & 2048 & 4096 & 8192 & 15360 & 32768\\
         \hline
         IN1k 10\% & SimCLR & 51.36 & 54.31 & 55.55 & 56.51 & 57.18 & \textbf{58.33}\\
          & VICReg & 54.72 & 57.96 & 59.13 & 59.88 & 60.38 & \textbf{60.45} \\
          & Byol & 43.11 & 45.94 & 47.71 & 47.95 & 50.61 & \textbf{52.37} \\
          & Supervised & \textbf{73.12} & 71.36 & 70.13 & 69.27 & 68.58 & 67.99 \\
        \hline
        CIFAR10 & SimCLR & 77.04 & 80.92 & 82.45 & 82.66 & 83.54 & \textbf{85.08} \\
        & VICReg & 76.76 & 81.09 & 83.2 & 85.25 & 85.77 & \textbf{86.7} \\ 
        & Byol & 68.76 & 72.62 & 75.21 & 75.32 & 78.41 & \textbf{78.71} \\
        & Supervised & 77.19 & 82.84 & 84.96 & 86.83 & 88.63 & \textbf{89.38} \\
        \hline
        CLEVR & SimCLR & 43.43 & 46.85 & 48.06 & 50.13 & 51.59 & \textbf{53.72} \\
        & VICReg & 42.12 & 44.75 & 46.96 & 48.03 & 50.92 & \textbf{52.75} \\
        & Byol & 36.41 & 38.56 & 40.35 & 40.47 & 45.44 & \textbf{47.93}\\
        & Supervised & 41.74 & 46.37 & 50.21 & 51.97 & 53.15 & \textbf{54.38} \\
        \hline
        Places & SimCLR & 46.62 & 50.77 & 52.59 & 53.61 & 54.63 & \textbf{55.54} \\
        & VICReg & 46.8 & 50.97 & 52.74  & 54.22 & 55.17 & \textbf{55.96} \\
        & Byol & 44.13 & 48.36 & 50.18 & 50.72 & 52.57 & \textbf{53.39} \\
        & Supervised & 47.98 & 51.9 & 53.53 & 54.3 & 55.09 & \textbf{55.65} \\
        \hline
        Eurosat & SimCLR & 87.26 & 88.52 & 88.2 & 89.96  & 90.16 & \textbf{91.68} \\
        & VICReg & 86.14 & 89.96 & 90.88 & 92.04 & 93.04 & \textbf{93.16} \\ 
        & Byol & 78.74 & 84.04 & 85.48 & 86.3 & 88.1 & \textbf{88.28} \\
        & Supervised & 84.46 & 88.82 & 91.54 & 93.64 & 94.96 & \textbf{95.92} \\
        \hline 
    \end{tabular}
    }
\end{table}

\paragraph{Nonlinear projector with balanced sampling}  To asses more broadly the impact of using wider representation in traditional SSL settings, we ran several experiment, using this time an uniform sampling strategy when training on ImageNet. We train SimCLR, VICReg, Byol and a supervised model on various backbone representation sizes while keeping the projector fixed\footnote{Experimental details can be found in the appendix.}. In \cref{fig:diff_acc_bin_cont}, we present the ImageNet top-1 validation accuracy for several methods across backbone dimensions $D$, using a linear probing evaluation. When using a traditional supervised training, the performances on ImageNet get better when $D$ becomes smaller. However, when looking at the SSL methods, there is a significant performance gain for larger $D$. This experiment highlights a clear limitation within the current SSL literature, which most often limits itself to using strictly the same backbone architectures that were developed for supervised learning. We hope that our study will encourage researchers to explore further SSL-specific \emph{architecture} designs enjoying the benefits of wider representations. 

Since increasing the size of the representation yields improved accuracy on ImageNet, we further \emph{evaluate} our pretrained models on a wider set of downstream tasks: ImageNet 10\% \citep{deng2009imagenet} (only 100 images are used for training), CIFAR10\citep{cifar10} , CLEVR \citep{Johnson2016CLEVRAD}, Places205\citep{place205}, Eurosat\citep{eurosat}. In \cref{tab:downstream_tasks}, we show that increasing the dimensionality of the backbone $D$ is beneficial for each of these datasets when using SSL methods. It is interesting to note that the best performance for a supervised model evaluated on ImageNet (in-distribution) are obtained when using small $D$, while the best performances when evaluating an ImageNet pretrained model on other datasets (out-of-distribution transfer) are obtained when using larger $D$. This confirms our insight that the ratio $D/K$ controls the amount of information and bias with respect to the pretraining task, i.e. the \emph{pretraining bias}. When training a supervised model on ImageNet and evaluating it on ImageNet, there is a perfect alignment between the tasks, thus performances are much better when using a small $D$ as the model is encouraged to learn invariances that will fit the task. However when training on ImageNet and evaluating on other datasets, it is beneficial to use a model pretrained on larger $D$, since the invariances learned with the pretraining objective might hurt the generalization on different downstream tasks.

\begin{figure*}[ht]
    \centering
    \includegraphics[scale=0.75]{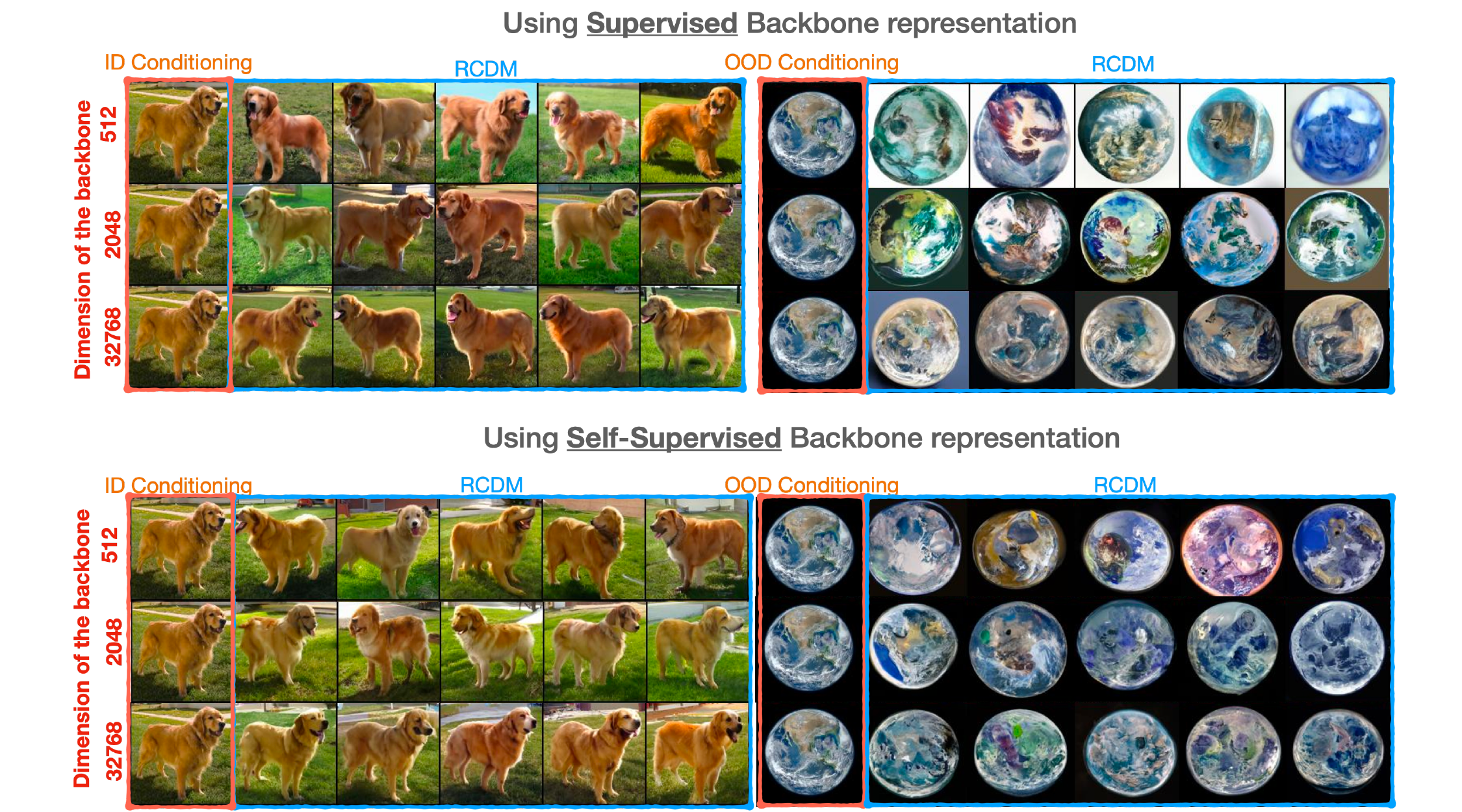}
    \caption{We use a conditional diffusion model (RCDM, \citet{RCDM}) trained on the face-blurred version of ImageNet \citep{yang2021imagenetfaces}, to visualize what information is retained in the representations given by models trained with different backbone representation size $D$. RCDM is conditioned on a representation and produces multiple samples. Aspects that remain constant across samples reflects information contained in the representation, while aspects that vary are not contained in the representation. We present two cases. In the first we condition on the representation of an in-distribution (ID) image,  taken from the validation set of ImageNet. In the second case we condition on an out-of-distribution (OOD) image from the Earth (Source: NASA). When considering the SSL model with $D=512$, the information that vary is the shadow and the pose of the animal (vertical flip) as well as some colors. However, when using SimCLR trained with $D=32768$, the shadow and animal pose remain much closer to the original input, meaning that increasing the dimension size results in retaining more information about a given input. We can see a similar trend in the OOD case: when the size of the representation is small, there is a higher variance in the generated sample which means that less information is encoded in the representation than in the case with higher representation size. The supervised model follows a similar trend except that the invariances are much stronger.} 
    \label{fig:RCDM_repr}
\end{figure*}

To verify this hypothesis, we run in \cref{fig:RCDM_repr} a qualitative visualization experiment using a representation conditional diffusion model (RCDM,  \citet{RCDM})  to map a given SSL representation back to image space. In this experiment we trained several RCDMs on the face-blurred version of ImageNet \citep{yang2021imagenetfaces} using the representations given by pretrained models having different backbone representation sizes. Then we use the representation of two images (one from the ImageNet validation set, the other from a public domain source) as conditioning for RCDM, and generate several samples using this conditioning. Aspects that remain constant across the samples is information that was retained in the representation, while aspects that vary across samples is information that is not contained in the representation. We observe that information about shadows or vertical flips doesn't seem to be present in the representation of SimCLR when using $D=512$. However, when looking at the largest backbone representation, RCDM is able to generate reconstructions much closer to the original image: retaining the pose, shape and similar color palette. Even in the out of distribution case, one can see that the information is better preserved when using the wider representation. In contrast, if we look at the representation learned with the supervised model, we can still see some invariances in the samples. What is also interesting is that when looking at the supervised model that is using $D=512$ for the backbone, there are a lot of invariances in the generated samples. For example the background of the dog is changing a bit while the drawing changes a lot. Such observation supports the hypothesis that having $D$ smaller than the number of classes might induces the learning of more invariances (since we decrease the amount of space in which the model can store information). This might also impacts robustness to some OOD settings since the model should learn to not rely on factors like background for classification. To verify such hypothesis, we ran an experiment on ImageNet-9 in \cref{tb:imagenet9} that show that indeed a representation that is smaller than the number of class is more robust to variation in the background. However, as showed in \cref{tab:downstream_tasks}, better performances and robustness on ImageNet don't necessarily translate to other downstream tasks.

\begin{table}[h]
    \centering
    \footnotesize
    \caption{\textbf{ImageNet-9} with a supervised backbone. In this experiment, we evaluate supervised models trained with various backbone dimension sizes on the ImageNet-9 benchmarks. As we observed in the RCDM experiment in \cref{fig:RCDM_repr}, the Resnet50 trained with the smaller backbone dimension is the most robust to perturbation of the background. 
    }
    \label{tb:imagenet9}
    \begin{tabular}{r|cccccc}
        \toprule
        Dim & original & only-fg & m-next & m-rand & m-same \\
        \toprule\toprule
        512 & \textbf{92.5} & \textbf{78.2} & \textbf{66.9} & \textbf{70.7} & \textbf{78.0}\\ 
        2048 & 91.7 & 74.2 & 64.3 & 66.8 & 75.8\\ 
        4096 & 91.8 & 72.2 & 63.4 & 65.2 & 74.8\\ 
        10240 & 90.1 & 71.5 & 61.3 & 63.7 & 73.4\\
        32768 & 90.0 & 70.0 & 59.7 & 62.5 & 71.7\\
    \end{tabular}
\end{table}

\begin{figure}[ht]
    \centering
    \includegraphics[scale=0.4]{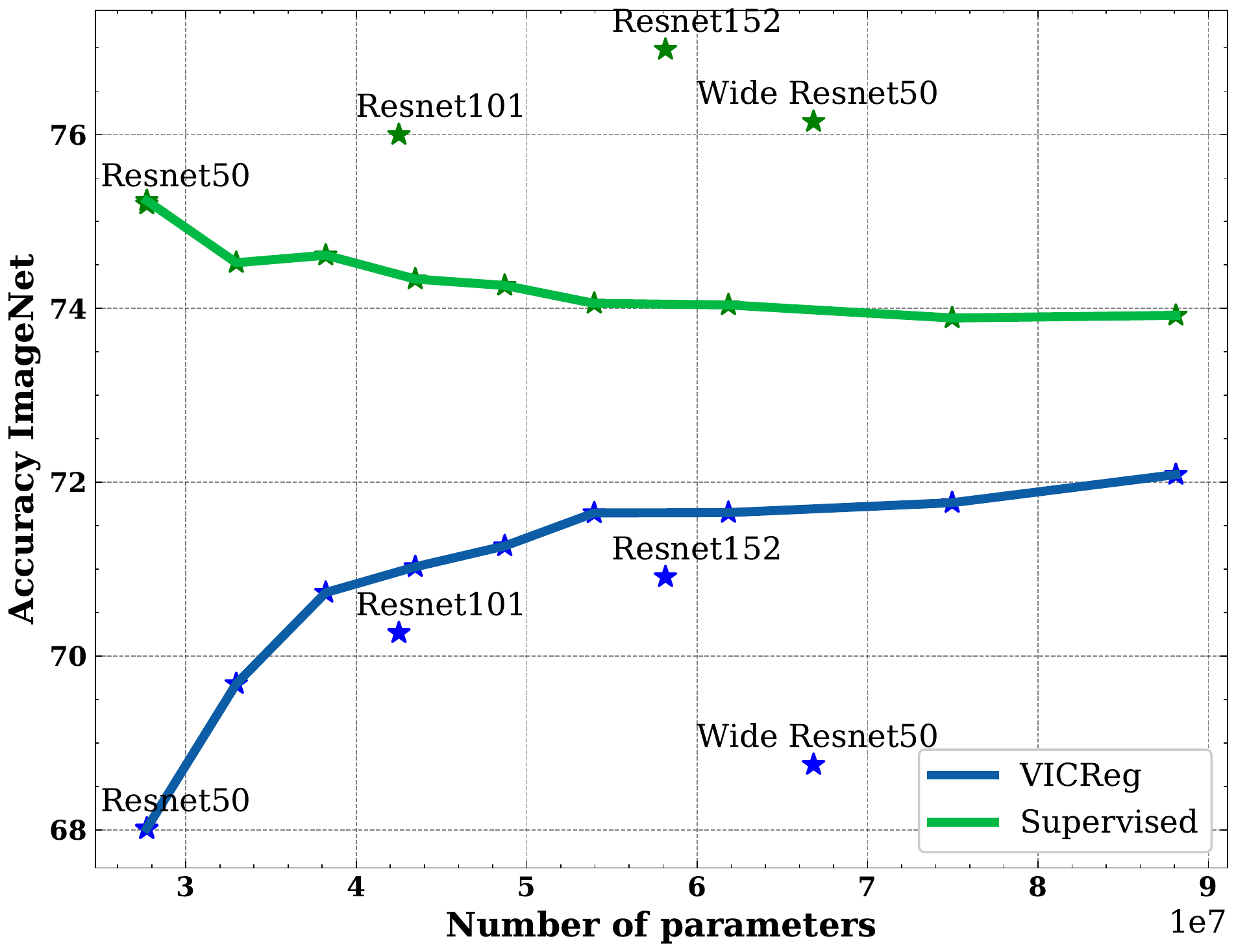}
    \caption{Top-1 accuracy on the validation set of imagenet with respect to the number of parameters in the model. The performances degrade with a supervised model when using wider representation whereas having deeper and wider network (while keeping the output backbone size fixed) help significantly. This is the opposite with a model like VIRCreg for which the best performances are obtained when using wider representation while deeper or wider network don't give as much gain in performances than in the supervised setting. This figure clearly show that what is working well for supervised training might not be the most suited for SSL.}
    \label{fig:params_vs_acc}
\end{figure}

\paragraph{Number of parameters} Increasing the number of features map in the last convolutional block of the Resnet increases significantly the number of parameters in the network. In consequence, there is still a question about how other Resnet variant, like deeper or wider network are located in this  picture. In \cref{fig:params_vs_acc}, we show the accuracy on the validation set of ImageNet with respect to the number of parameters in the model. Many Resnet that were developed to boost the performances in Supervised training don't work as well with SSL. A Resnet50 with wider backbone representation beat the deepest and wider Resnet when using VICReg.

\section{Wider representations are sparse, more linearly separable and binarizable}

\subsection{Wider SSL representations are sparse}

\begin{figure}[ht]
    \centering
    \includegraphics[scale=0.32]{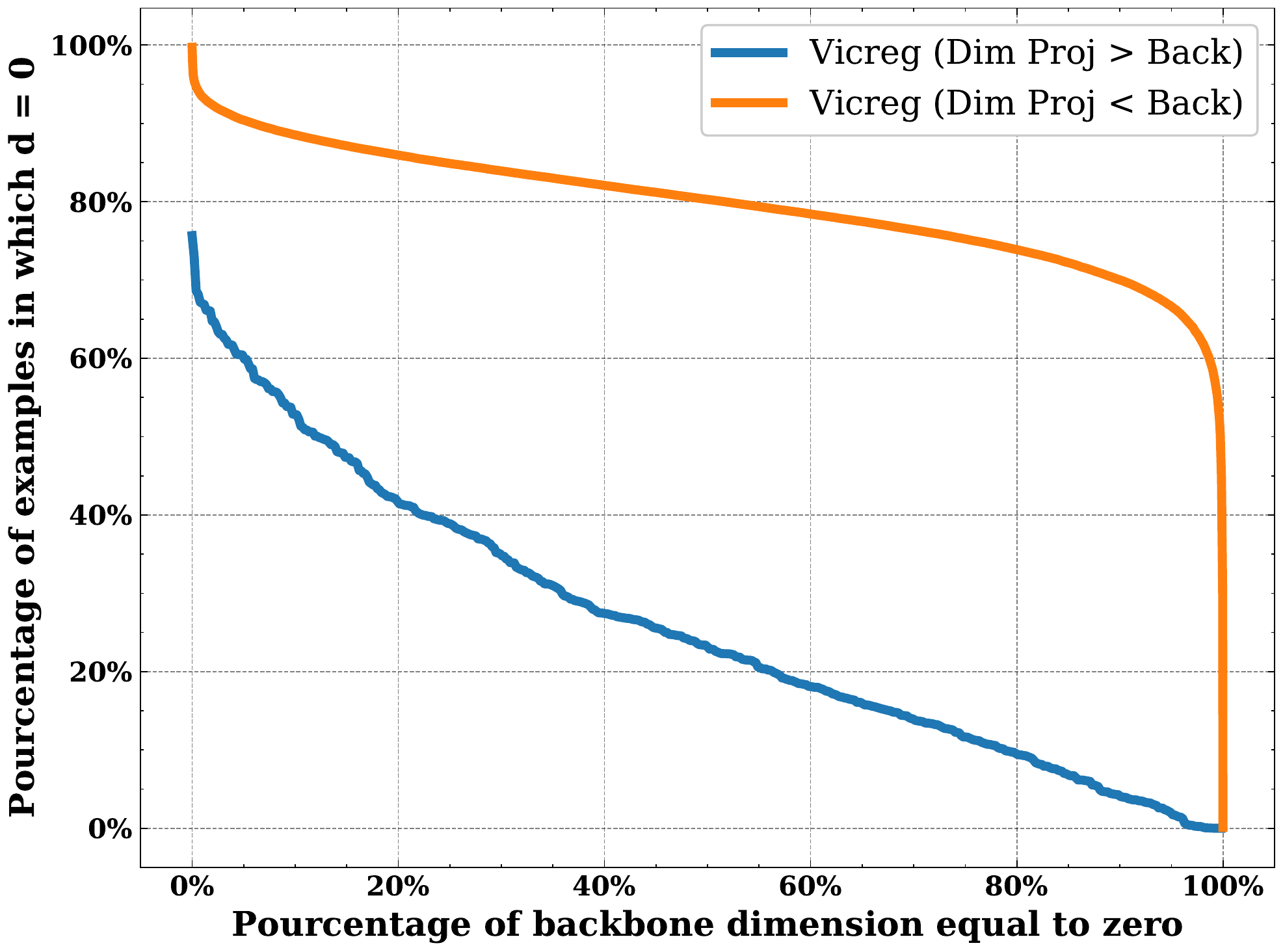}
    \caption{In this experiment we fixed the dimension of the projector to 8192 and change the backbone dimension to 512 for the first experiment and to 32768 for the second experiment. Then we trained VicReg for each of these two setups and study how many times a given dimension in the backbone space is equal to zero over the entire validation set.} 
    \label{fig:visu_sparsity}
\end{figure}

We validate the hypothesis that using a larger ratio $D/K$ results in learning a representation that spans only a subset of the backbone dimension when using a nonlinear projector by measuring the sparsity for each example in the validation set of ImageNet. We performed two experiments using  VICReg. In the first experiment, we use a backbone of lower dimension than the projector's ($K=8192,~D=512$). In the second, we use a backbone of higher dimension than the projector's ($K=8192,~D=32768$). We evaluate these models by computing the number of activations equal to zero in the representation\footnote{The ReLU activation at the backbone's output induces the sparsity.}. Then, we sort these values and present them in~\cref{fig:visu_sparsity}. We observe that the amount of activations that are $0$ is very low when the backbone's dimension is lower than the projector's. Thus, most of the examples will span the entire backbone vector. In contrast, we observe a very sparse bacbone's output when $D$ is larger than $K$ with 80\% of the examples that have half of the backbone representation's activation equal to zero.

This experiment highlights that in addition to learning more information about the data when using larger backbone dimensions, the representation learned are more sparse. 

\subsection{Wider representation are more easily linearly separable} 
\begin{figure}
    \centering
    \includegraphics[scale=0.4]{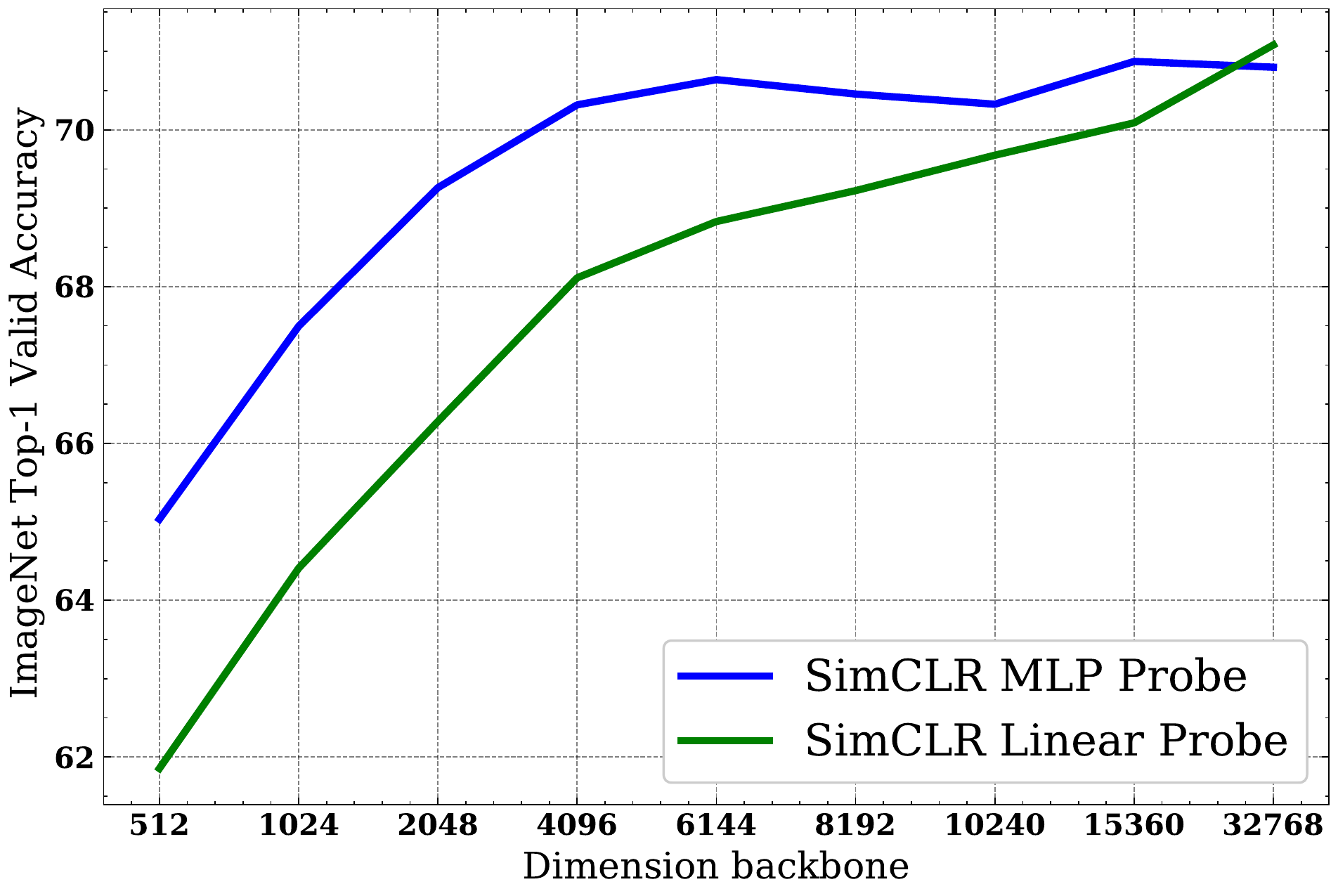}
    \caption{In this experiment, we compare the performances on the validation set of ImageNet when using a linear evaluation versus using a nonlinear one (with a 3 layer MLP) on pretrained SimCLR models trained with a varying backbone dimension $D$. When using smaller representations the gap is large and becomes much smaller with wider representations.}
    \label{fig:linear_vs_mlp}
\end{figure}

Many SSL works use the linear probing as a way to evaluate their model, however such evaluation protocol can be limited if the information is entangled. In consequence, it's important when evaluating a SSL model to compare the performances using a linear and a nonlinear probe. In \cref{fig:linear_vs_mlp}, we comapre the performances using those two evaluations methods for a varying backbone dimension $D$. We show for SimCLR that there is an important gap between the performances one might get on ImageNet with a simple linear probing versus a nonlinear one when the $D$ is small. However when looking at higher dimension the differences between both methods decrease significantly. This mostly implies that the features are more linearly separable when using wider representation vectors.

\subsection{Wider representations are binarizable}
\label{sec:good}

\begin{figure}
    \centering
    \includegraphics[scale=0.36]{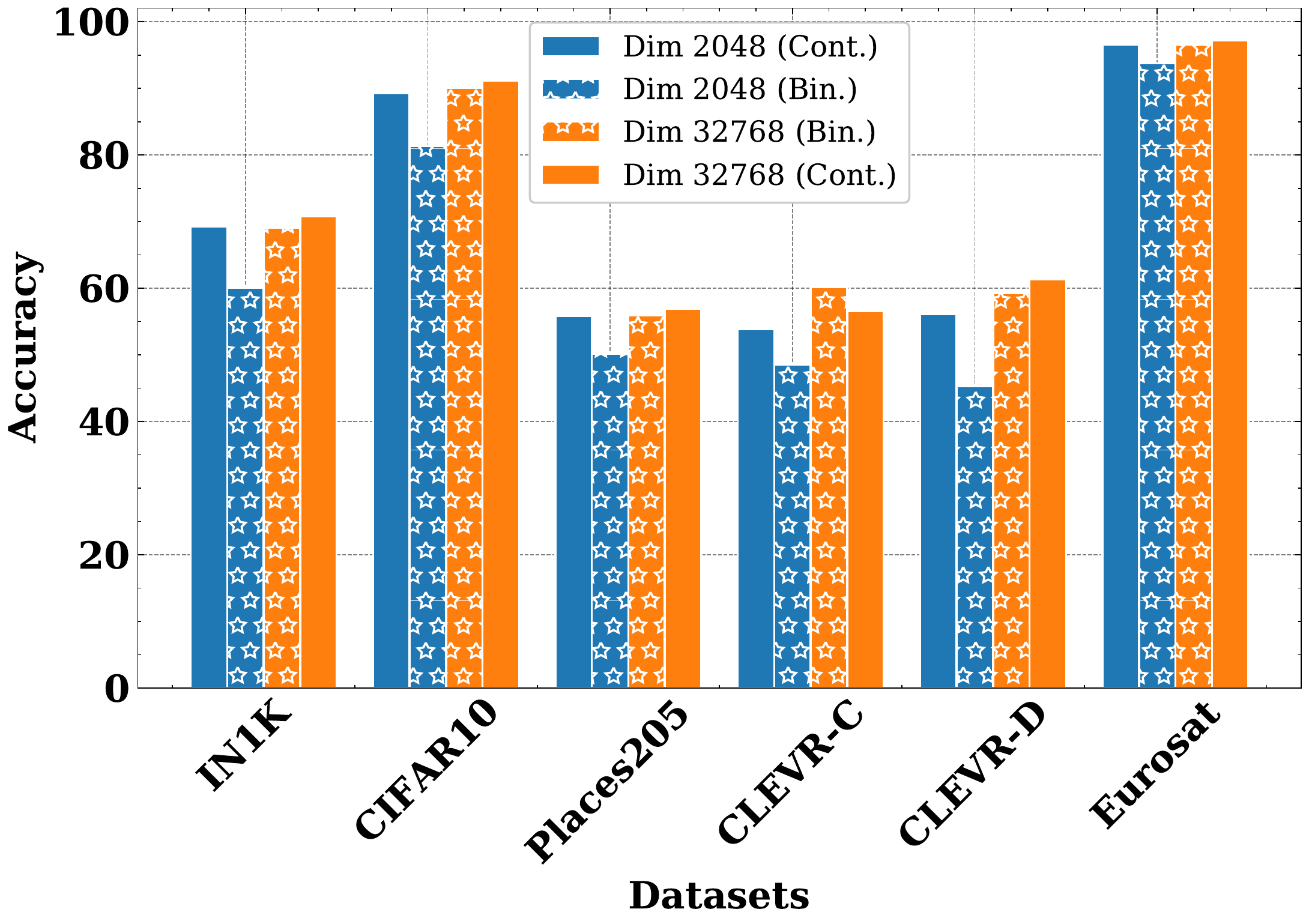}
    \caption{Transfer learning with SimCLR on various downstream tasks for binary and continuous representation. We observe on ImageNet-1k (IN1K) that we achieve similar performances when using a representation size of 32768 with continuous and binary representation while there is a clear drop in performances when using a backbone representations of size 2048. This plot shows how much wider representation can be binarized without loss of performance. }
    \label{fig:dataset_bin}
\end{figure}

Since wider representation are more sparse, we evaluate how easily they are binarizable. The intuition is that if one has roughly symmetrical, centered and independent distributions between the embeddings dimensions, then performing quantization will only collapse different images to the same quantized code with very low probability --going exponentially quickly to $0$ with the embedding dimension (demonstrated in Appendix \ref{sec:proof}). Since binarization should occur naturally when using wider embedding, we ran an experiment in which we measure the performances on a SimCLR model trained with a backbone representation size of 2048 and another one trained with a size of 32768 and compare their performances on different downstream tasks when using a continuous reprensentation and a binarized ones. The binarization operation is simply performed by settings all non zeros elements in the representation to one. In \cref{fig:dataset_bin}, we show that when using a representation size of 2048, the performances decrease significantly when binarizing the representation. In contrast, when using a wider representation, the performances between the binarized and continuous representation are extremely close. Having the ability to save binarized version of the representation might alleviate the additional memory cost in storing the representation due to the use of larger vectors.

\section{Conclusion}
In this work, we studied the transfer effectiveness of a very simple trick, which consists in controlling in the simplest possible way the dimension of the backbone representation learned during pretraining. When using a linear projector, having a large backbone and a small projector improves the performance on both balanced and unbalanced data. When using a nonlinear projector, significant gains can also be achieved for SSL methods by only expanding the backbone dimension while keeping the projector fixed. In a supervised setting, we shed light on a strong pretraining bias phenomenon: performance of models are higher when using smaller backbone if the downstream task is the same as the pretraining task, but when performing transfer it is preferable to use larger representations. Lastly, we show that larger representations are binarizable without any significant performance loss. We hope that these insights will guide SSL practitioners in designing better architectures for SSL methods, as well as better transfer learning techniques.
\paragraph{Limitations and future works} This paper focuses its attention on ResNet architectures. While the study of vision transformers is also relevant, increasing their backbone dimensionality requires more technical considerations that we leave as future works.

\nocite{langley00}

\bibliography{main}

\begin{thebibliography}{27}
\providecommand{\natexlab}[1]{#1}
\providecommand{\url}[1]{\texttt{#1}}
\expandafter\ifx\csname urlstyle\endcsname\relax
  \providecommand{\doi}[1]{doi: #1}\else
  \providecommand{\doi}{doi: \begingroup \urlstyle{rm}\Url}\fi

\bibitem[Assran et~al.(2022{\natexlab{a}})Assran, Balestriero, Duval, Bordes,
  Misra, Bojanowski, Vincent, Rabbat, and Ballas]{uniform_prior}
Assran, M., Balestriero, R., Duval, Q., Bordes, F., Misra, I., Bojanowski, P.,
  Vincent, P., Rabbat, M., and Ballas, N.
\newblock The hidden uniform cluster prior in self-supervised learning,
  2022{\natexlab{a}}.
\newblock URL \url{https://arxiv.org/abs/2210.07277}.

\bibitem[Assran et~al.(2022{\natexlab{b}})Assran, Caron, Misra, Bojanowski,
  Bordes, Vincent, Joulin, Rabbat, and Ballas]{msn}
Assran, M., Caron, M., Misra, I., Bojanowski, P., Bordes, F., Vincent, P.,
  Joulin, A., Rabbat, M., and Ballas, N.
\newblock Masked siamese networks for label-efficient learning.
\newblock In \emph{Computer Vision – ECCV 2022: 17th European Conference, Tel
  Aviv, Israel, October 23–27, 2022, Proceedings, Part XXXI}, pp.\
  456–473, Berlin, Heidelberg, 2022{\natexlab{b}}. Springer-Verlag.
\newblock ISBN 978-3-031-19820-5.
\newblock \doi{10.1007/978-3-031-19821-2_26}.
\newblock URL \url{https://doi.org/10.1007/978-3-031-19821-2_26}.

\bibitem[Bordes et~al.(2021)Bordes, Balestriero, and Vincent]{RCDM}
Bordes, F., Balestriero, R., and Vincent, P.
\newblock High fidelity visualization of what your self-supervised
  representation knows about.
\newblock \emph{CoRR}, abs/2112.09164, 2021.
\newblock URL \url{https://arxiv.org/abs/2112.09164}.

\bibitem[Bordes et~al.(2022)Bordes, Balestriero, Garrido, Bardes, and
  Vincent]{Guillotine}
Bordes, F., Balestriero, R., Garrido, Q., Bardes, A., and Vincent, P.
\newblock Guillotine regularization: Improving deep networks generalization by
  removing their head, 2022.
\newblock URL \url{https://arxiv.org/abs/2206.13378}.

\bibitem[Caron et~al.(2020)Caron, Misra, Mairal, Goyal, Bojanowski, and
  Joulin]{caron2020swav}
Caron, M., Misra, I., Mairal, J., Goyal, P., Bojanowski, P., and Joulin, A.
\newblock Unsupervised learning of visual features by contrasting cluster
  assignments.
\newblock In \emph{NeurIPS}, 2020.

\bibitem[Chen et~al.(2020)Chen, Kornblith, Norouzi, and Hinton]{chen2020simclr}
Chen, T., Kornblith, S., Norouzi, M., and Hinton, G.~E.
\newblock A simple framework for contrastive learning of visual
  representations.
\newblock In \emph{ICML}, 2020.

\bibitem[Deng et~al.(2009)Deng, Dong, Socher, Li, Li, and
  Fei-Fei]{deng2009imagenet}
Deng, J., Dong, W., Socher, R., Li, L.-J., Li, K., and Fei-Fei, L.
\newblock Imagenet: A large-scale hierarchical image database.
\newblock In \emph{CVPR}, 2009.

\bibitem[Dubois et~al.(2022)Dubois, Hashimoto, Ermon, and
  Liang]{dubois_improving_2022}
Dubois, Y., Hashimoto, T., Ermon, S., and Liang, P.
\newblock Improving {Self}-{Supervised} {Learning} by {Characterizing}
  {Idealized} {Representations}, December 2022.
\newblock URL \url{http://arxiv.org/abs/2209.06235}.
\newblock arXiv:2209.06235 [cs, stat].

\bibitem[Garrido et~al.(2022)Garrido, Chen, Bardes, Najman, and
  Lecun]{garrido_duality_2022}
Garrido, Q., Chen, Y., Bardes, A., Najman, L., and Lecun, Y.
\newblock On the duality between contrastive and non-contrastive
  self-supervised learning, October 2022.
\newblock URL \url{http://arxiv.org/abs/2206.02574}.
\newblock arXiv:2206.02574 [cs].

\bibitem[Gidaris et~al.(2018)Gidaris, Singh, and Komodakis]{Gidaris2018arxiv}
Gidaris, S., Singh, P., and Komodakis, N.
\newblock Unsupervised representation learning by predicting image rotations.
\newblock In \emph{6th International Conference on Learning Representations,
  {ICLR} 2018, Vancouver, BC, Canada, April 30 - May 3, 2018, Conference Track
  Proceedings}. OpenReview.net, 2018.
\newblock URL \url{https://openreview.net/forum?id=S1v4N2l0-}.

\bibitem[He et~al.(2016)He, Zhang, Ren, and Sun]{he2016resnet}
He, K., Zhang, X., Ren, S., and Sun, J.
\newblock Deep residual learning for image recognition.
\newblock In \emph{CVPR}, 2016.

\bibitem[He et~al.(2022)He, Chen, Xie, Li, Doll{\'a}r, and
  Girshick]{he2022masked}
He, K., Chen, X., Xie, S., Li, Y., Doll{\'a}r, P., and Girshick, R.
\newblock Masked autoencoders are scalable vision learners.
\newblock In \emph{Proceedings of the IEEE/CVF Conference on Computer Vision
  and Pattern Recognition}, pp.\  16000--16009, 2022.

\bibitem[Helber et~al.(2017)Helber, Bischke, Dengel, and Borth]{eurosat}
Helber, P., Bischke, B., Dengel, A., and Borth, D.
\newblock Eurosat: A novel dataset and deep learning benchmark for land use and
  land cover classification.
\newblock \emph{CoRR}, abs/1709.00029, 2017.
\newblock URL
  \url{http://dblp.uni-trier.de/db/journals/corr/corr1709.html#abs-1709-00029}.

\bibitem[Horn et~al.(2018)Horn, Aodha, Song, Cui, Sun, Shepard, Adam, Perona,
  and Belongie]{inat18}
Horn, G.~V., Aodha, O.~M., Song, Y., Cui, Y., Sun, C., Shepard, A., Adam, H.,
  Perona, P., and Belongie, S.~J.
\newblock The inaturalist species classification and detection dataset.
\newblock In \emph{CVPR}, pp.\  8769--8778. IEEE Computer Society, 2018.
\newblock URL
  \url{http://dblp.uni-trier.de/db/conf/cvpr/cvpr2018.html#HornASCSSAPB18}.

\bibitem[Johnson et~al.(2016)Johnson, Hariharan, van~der Maaten, Fei-Fei,
  Zitnick, and Girshick]{Johnson2016CLEVRAD}
Johnson, J., Hariharan, B., van~der Maaten, L., Fei-Fei, L., Zitnick, C.~L.,
  and Girshick, R.~B.
\newblock Clevr: A diagnostic dataset for compositional language and elementary
  visual reasoning.
\newblock \emph{2017 IEEE Conference on Computer Vision and Pattern Recognition
  (CVPR)}, pp.\  1988--1997, 2016.

\bibitem[Krizhevsky(2009)]{cifar10}
Krizhevsky, A.
\newblock Learning multiple layers of features from tiny images.
\newblock pp.\  32--33, 2009.
\newblock URL
  \url{https://www.cs.toronto.edu/~kriz/learning-features-2009-TR.pdf}.

\bibitem[Loshchilov \& Hutter(2019)Loshchilov and Hutter]{adamw}
Loshchilov, I. and Hutter, F.
\newblock Decoupled weight decay regularization.
\newblock In \emph{7th International Conference on Learning Representations,
  {ICLR} 2019, New Orleans, LA, USA, May 6-9, 2019}. OpenReview.net, 2019.
\newblock URL \url{https://openreview.net/forum?id=Bkg6RiCqY7}.

\bibitem[Noroozi \& Favaro(2016)Noroozi and Favaro]{Noroozi_2016_Jigsaw}
Noroozi, M. and Favaro, P.
\newblock Unsupervised learning of visual representations by solving jigsaw
  puzzles.
\newblock In \emph{Proceedings of the European Conference on Computer Vision
  (ECCV)}, 2016.

\bibitem[Olshausen \& Field(1997)Olshausen and Field]{sparse_coding}
Olshausen, B.~A. and Field, D.~J.
\newblock Sparse coding with an overcomplete basis set: A strategy employed by
  v1?
\newblock \emph{Vision Research}, 37\penalty0 (23):\penalty0 3311--3325, 1997.
\newblock ISSN 0042-6989.
\newblock \doi{https://doi.org/10.1016/S0042-6989(97)00169-7}.
\newblock URL
  \url{https://www.sciencedirect.com/science/article/pii/S0042698997001697}.

\bibitem[Ranzato et~al.(2006)Ranzato, Poultney, Chopra, and Cun]{yann_sparsity}
Ranzato, M.~a., Poultney, C., Chopra, S., and Cun, Y.
\newblock Efficient learning of sparse representations with an energy-based
  model.
\newblock In Sch\"{o}lkopf, B., Platt, J., and Hoffman, T. (eds.),
  \emph{Advances in Neural Information Processing Systems}, volume~19. MIT
  Press, 2006.
\newblock URL
  \url{https://proceedings.neurips.cc/paper/2006/file/87f4d79e36d68c3031ccf6c55e9bbd39-Paper.pdf}.

\bibitem[Tian et~al.(2021)Tian, Henaff, and van~den Oord]{tian2021divide}
Tian, Y., Henaff, O.~J., and van~den Oord, A.
\newblock Divide and contrast: Self-supervised learning from uncurated data.
\newblock In \emph{Proceedings of the IEEE/CVF International Conference on
  Computer Vision}, pp.\  10063--10074, 2021.

\bibitem[Vincent et~al.(2008)Vincent, Larochelle, Bengio, and
  Manzagol]{denoising_vincent}
Vincent, P., Larochelle, H., Bengio, Y., and Manzagol, P.-A.
\newblock Extracting and composing robust features with denoising autoencoders.
\newblock In \emph{Proceedings of the 25th International Conference on Machine
  Learning}, ICML '08, pp.\  1096–1103, New York, NY, USA, 2008. Association
  for Computing Machinery.
\newblock ISBN 9781605582054.
\newblock \doi{10.1145/1390156.1390294}.
\newblock URL \url{https://doi.org/10.1145/1390156.1390294}.

\bibitem[Wang \& Isola(2020)Wang and Isola]{wang2020hypersphere}
Wang, T. and Isola, P.
\newblock Understanding contrastive representation learning through alignment
  and uniformity on the hypersphere.
\newblock In \emph{International Conference on Machine Learning}, pp.\
  9929--9939. PMLR, 2020.

\bibitem[Yang et~al.()Yang, Yau, Fei-Fei, Deng, and
  Russakovsky]{yang2021imagenetfaces}
Yang, K., Yau, J., Fei-Fei, L., Deng, J., and Russakovsky, O.
\newblock A study of face obfuscation in imagenet.
\newblock In \emph{International Conference on Machine Learning (ICML)}.

\bibitem[You et~al.(2017)You, Gitman, and Ginsburg]{you2017lars}
You, Y., Gitman, I., and Ginsburg, B.
\newblock Large batch training of convolutional networks.
\newblock \emph{arXiv preprint arXiv:1708.03888}, 2017.

\bibitem[Zbontar et~al.(2021)Zbontar, Jing, Misra, LeCun, and
  Deny]{zbontar2021barlow}
Zbontar, J., Jing, L., Misra, I., LeCun, Y., and Deny, S.
\newblock Barlow twins: Self-supervised learning via redundancy reduction.
\newblock \emph{arXiv preprint arxiv:2103.03230}, 2021.

\bibitem[Zhou et~al.(2014)Zhou, Lapedriza, Xiao, Torralba, and Oliva]{place205}
Zhou, B., Lapedriza, A., Xiao, J., Torralba, A., and Oliva, A.
\newblock Learning deep features for scene recognition using places database.
\newblock In Ghahramani, Z., Welling, M., Cortes, C., Lawrence, N., and
  Weinberger, K. (eds.), \emph{Advances in Neural Information Processing
  Systems}, volume~27. Curran Associates, Inc., 2014.
\newblock URL
  \url{https://proceedings.neurips.cc/paper/2014/file/3fe94a002317b5f9259f82690aeea4cd-Paper.pdf}.

\end{thebibliography}
\bibliographystyle{icml2022}

\newpage
\appendix
\onecolumn

\section{Experimental details.}

\begin{figure}[ht]
    \centering
    \includegraphics[scale=0.3]{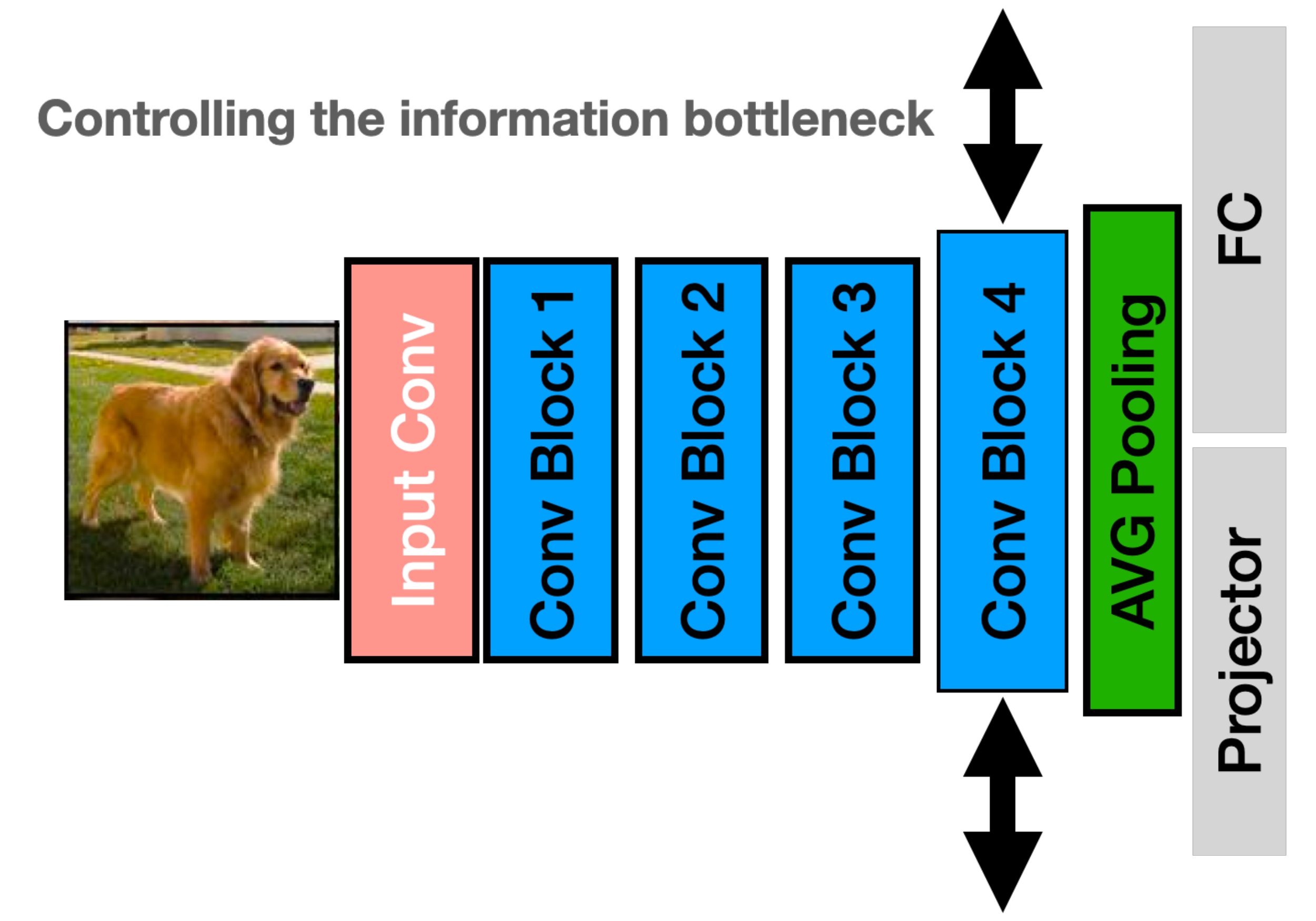}
    \caption{We control the information bottleneck by decreasing or increasing the number of features map at the last convolutional block of a Resnet50. Since this is the last layer before the pooling operation, increasing the number of features maps allow us to retain more information across the network. Traditionally, Resnet50 have used 2048 features map which is the common setup when training such method in supervised and self-supervised learning. In our experiments, we vary the number of features map from 512 to 32768. } 
    \label{fig:control_schema}
\end{figure}

In this paper, we focus our work on Resnet50 \citep{he2016resnet} in which we change the number of features in its last residual block. This operation is performed before pooling in order for the network to be able to retain as much information possible about its input. This is illustrated in \cref{fig:control_schema}. We train all our SSL models on ImageNet \citep{deng2009imagenet} using a batch size of $512$ and a projector of size $8192-8192-8192$ with Relu activation and batch normalization unless specified otherwise. SimCLR is trained with a temperature of $0.15$ using the LARS \citep{you2017lars} optimizer and a cyclic cosine learning rate whose base value is $0.5$ with a weight decay set to $1e-6$. VICReg is trained with a similarity and standard deviation coefficient of 25 while using a covariance coefficient of 1. We also used LARS as optimizer with the same learning rates as SimCLR except that we set its weight decay to $1e-4$. The supervised model is also trained with LARS as optimizer using the same learning rate as the SSL methods and with a weight decay of $1e-6$. All models were trained on 100 epochs unless specified otherwise and were trained using data augmentations: RandromCropping, HorizontalFlip, ColorJittering, Grayscaling and Solarization. Concerning RCDM, we have used the same parameters as in \citet{RCDM}. When evaluating the models, we use a linear prob on top of the backbone representation that we trained for 100 epochs with the AdamW \citep{adamw} optimizer, a learning rate of $1e-4$ and weight decay of $0.04$ with a batch size of $1024$.

\section{Additional experiments}

\begin{figure}[ht]
    \centering
    \includegraphics[scale=0.4]{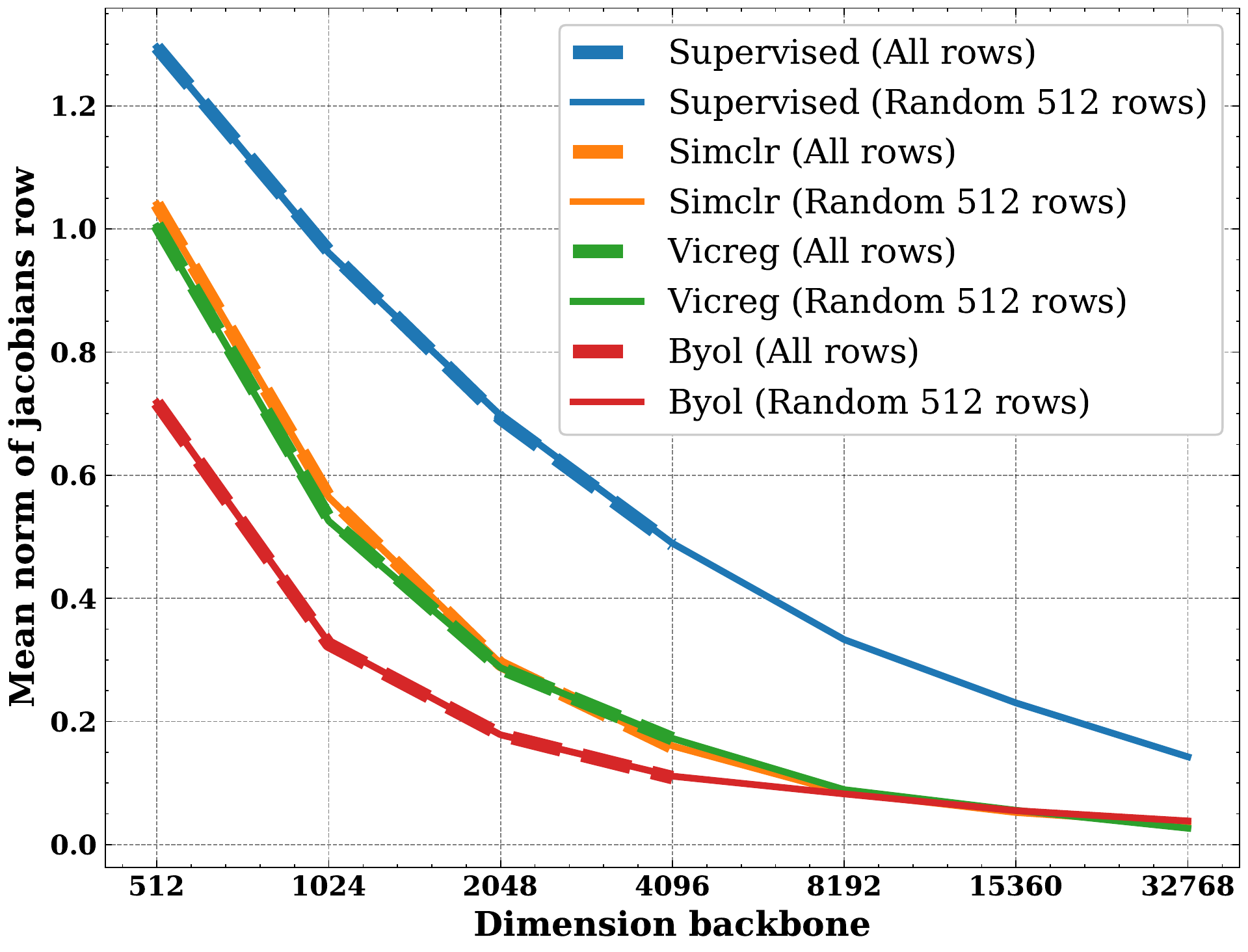}
    \caption{Wwe compute the mean norm of the jacobian for each individual dimension in the backbone representation. The intuition behind this figure is to have some proxy to get insights on how much information is stored in a single individual dimension by measuring the sensitivity of this dimensions with respect to small inputs' perturbations. We see that increasing $D$ leads to lower norm per dimension for SSL and supervised methods. } 
    \label{fig:jacobian}
\end{figure}

\paragraph{Norm of the jacobian} To asses how much increasing the backbone dimension decreases the amount of information in a given dimension, we compute the norm of the Jacobian matrix for each individual backbone dimension and compute the mean across all the dimensions. In \cref{fig:jacobian}, we plot the mean norm of the jacobian for different pretraining $D$ and different type of SSL criteria. For both Supervised and SSL models, increasing $D$ lead to significant smaller norm for the jacobian which imply that each of these dimensions will become more robust to slight changes in their inputs. 

\begin{table}[h]
    \centering
    \footnotesize
    \caption{\textbf{Different projector architecture} when training SimCLR on ImageNet. In these experiments, we kept the size of the backbone fixed to 32768 and change only the architecture of the projector.  
    }
    \label{tb:architecture_proj}
    \begin{tabular}{|c|c|c|c|}
        \toprule
        No-Proj & Linear (32) & Linear (8192) & Linear (16834)\\
        46.7 & 67.4 & 65.5 & 65.3 \\
        \hline
        nonlinear (128-128-128) & nonlinear (32-8192) & nonlinear (32-8192-8192) & nonlinear (8192-8192-32)  \\
        65.1 & 67.4 & 68.9 & 70.1 \\
        \toprule
    \end{tabular}
\end{table}

\begin{figure}[ht]
    \centering
    \includegraphics[scale=0.3]{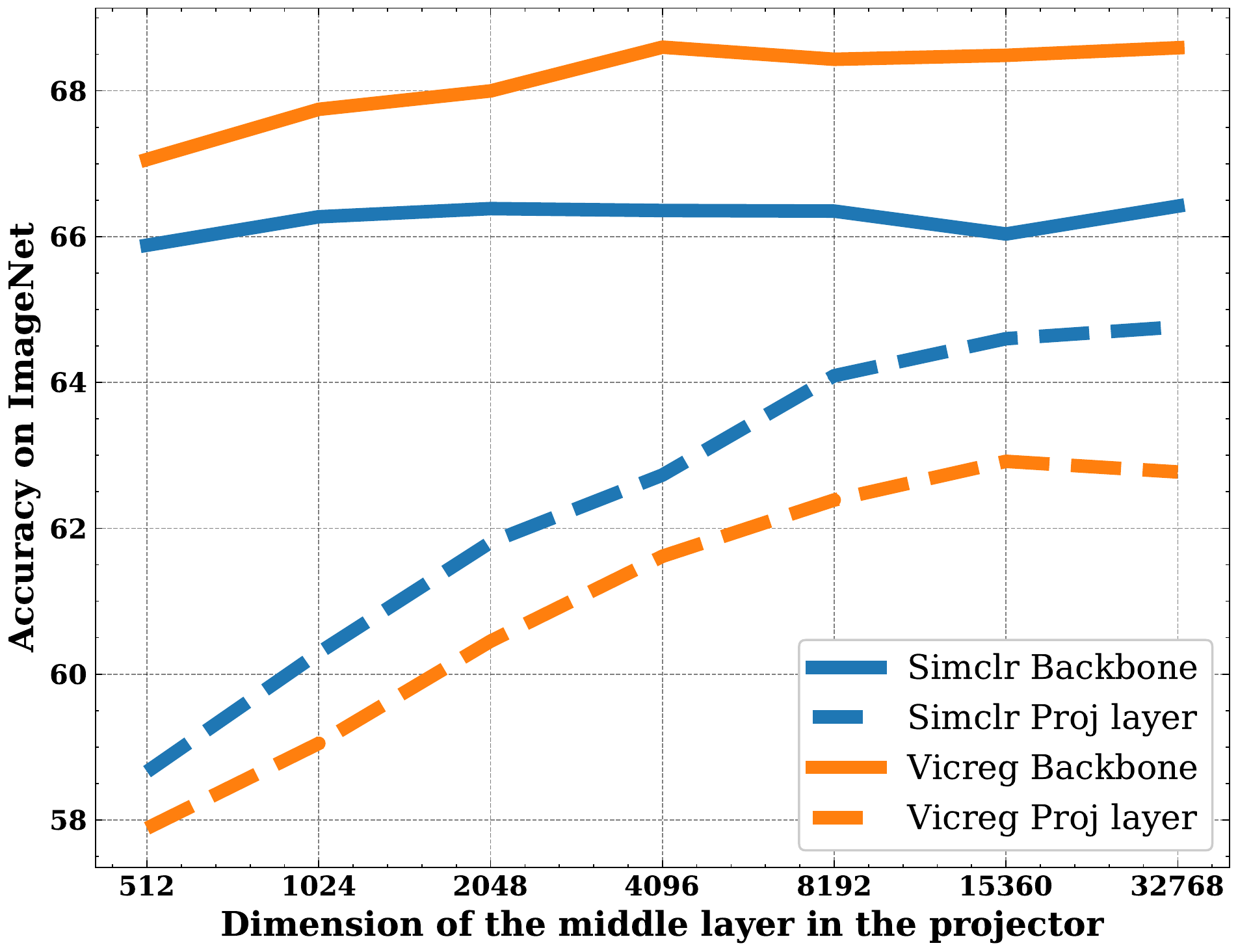}
    \caption{Linear probing accuracy on the validation set of ImageNet using VicReg and SimCLR pretrained on different middle projector layer dimension $K$ while keeping the first and last dimension fixed to $8192$). We observe that having a larger middle layer improves a bit the accuracy with VICReg, however the improvement has a much lower magnitude than the ones we got when increasing the backbone representation. } 
    \label{fig:middle_layer_proj}
\end{figure}

\paragraph{Changing the projector architecture} Another experiment we performed was to change only the middle layer dimension of the projector when training SimCLR and VICReg on ImageNet. Doing this does not produce a strong impact at the backbone level as shown in \cref{fig:middle_layer_proj}. However when training a linear probe on this middle layer, on can see performances improving (despite being still bellow the performances at the backbone level). We extend our experiment by trying different dimension for the first and last layer of the projector in \cref{tb:architecture_proj}. It's interesting to note that we get better performances when using a very small linear projector (32 units) than when using a nonlinear projector with 3 layers of size 128. However, the best performances are still obtained with wider first projector layer representation in the nonlinear case.

\paragraph{Extended downstream tasks evaluations} 
We present in \cref{tab:downstream_tasks_all} a more complete evaluation of our models on different downsteam tasks.

\begin{figure}[t!]
    \centering
    \footnotesize
    \begin{minipage}{0.33\linewidth}
    \centering
    \includegraphics[scale=0.4]{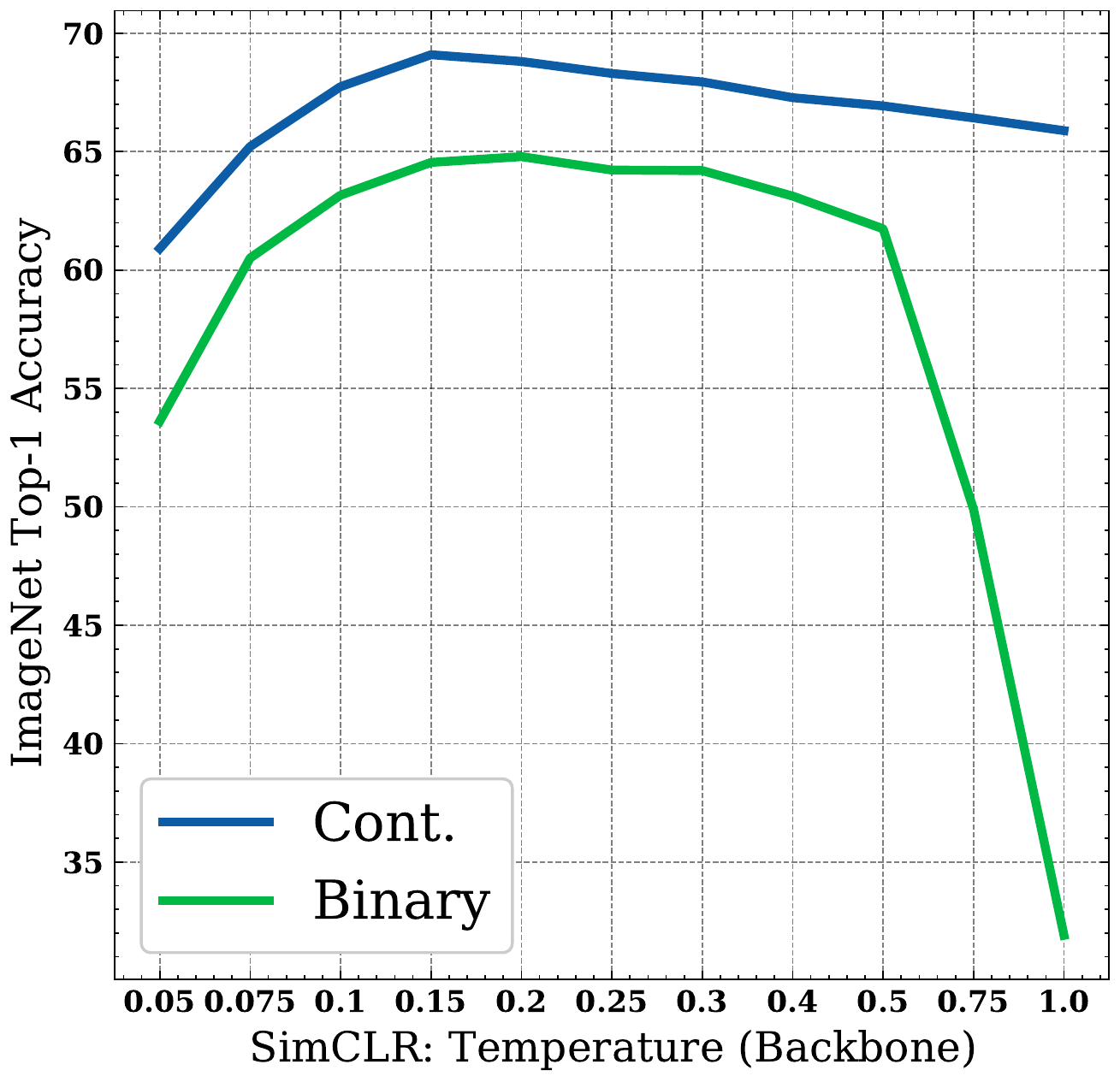}
    \label{fig:simclr_supervised}
    \subcaption{Backbone}
    \end{minipage}%
    \begin{minipage}{0.33\linewidth}
    \centering
    \includegraphics[scale=0.4]{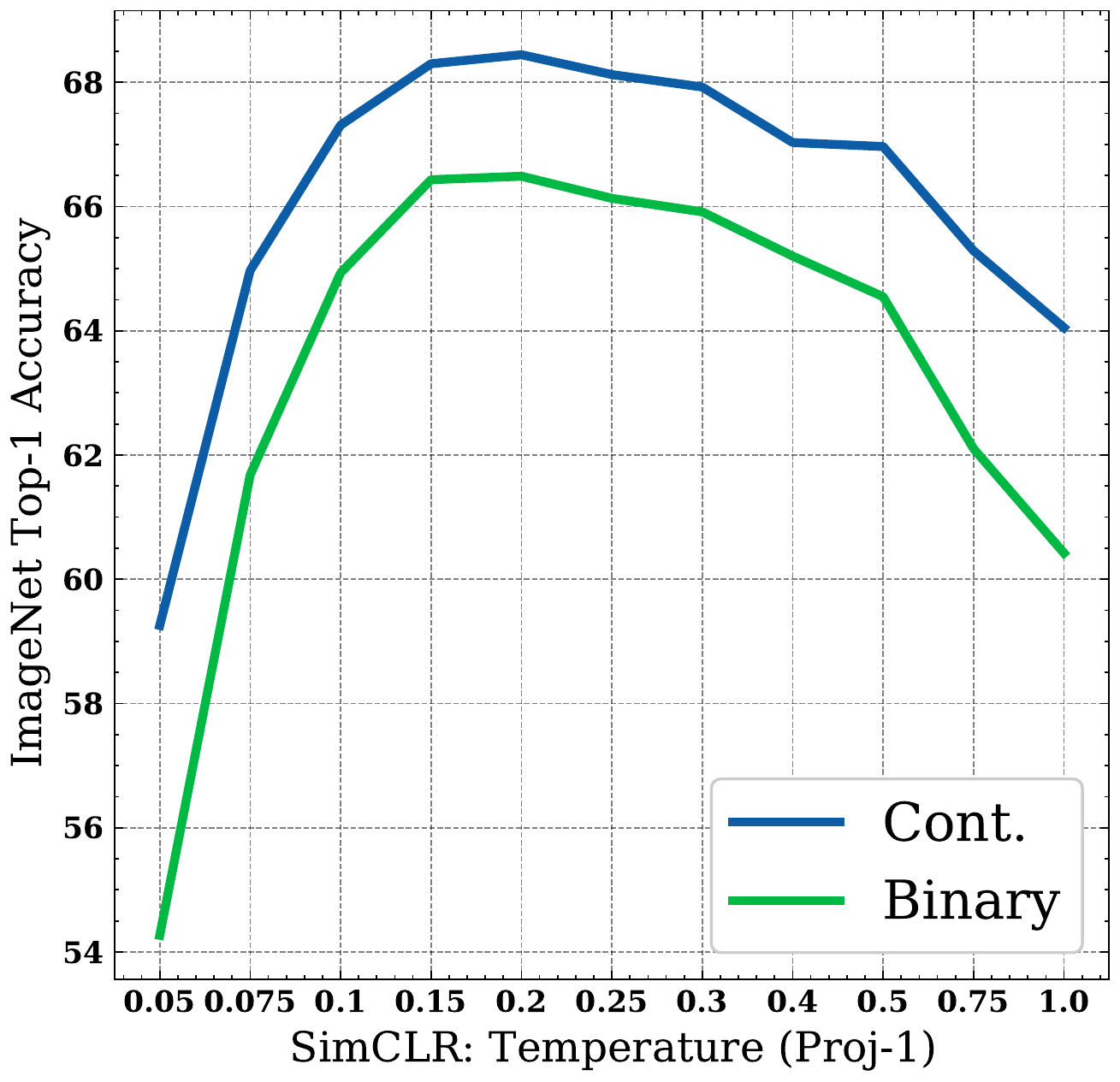}
    \label{fig:simclr_supervised}
    \subcaption{Proj-1}
    \end{minipage}%
    \begin{minipage}{0.33\linewidth}
    \centering
    \includegraphics[scale=0.4]{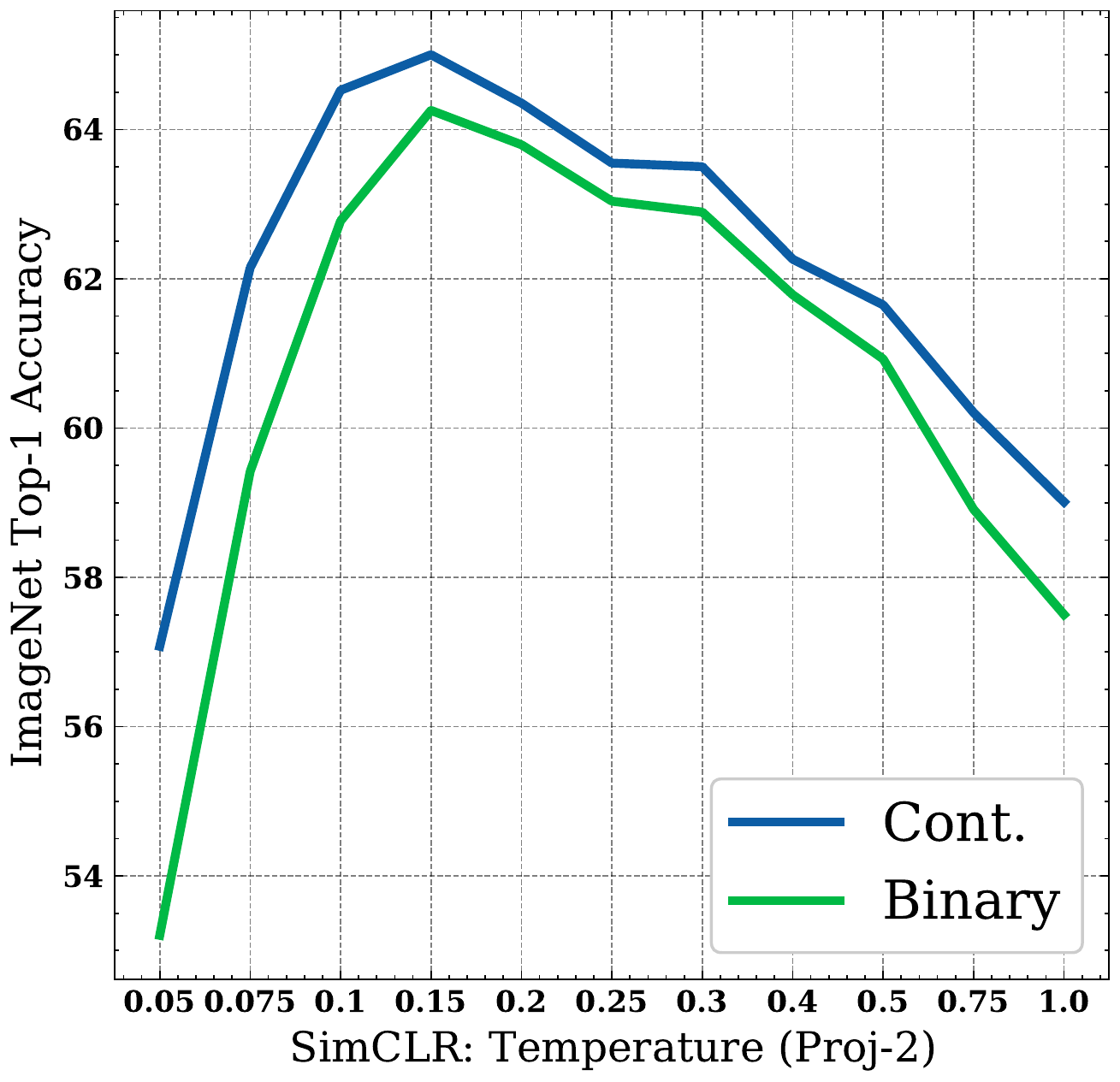}
    \label{fig:exp_supervised_co3d}
    \subcaption{Proj-2}
    \end{minipage}
    \caption{Binarization robustness with respect to the SimCLR temperature. In this experiment we train SimCLR with various values as temperature for a fixed backbone size $D=10240$ and projector size $8192-8192-8192$. Then, we train and compute with a linear probe the top-1 imagenet validation accuracy. It's interesting to note that using higher temperature with SimCLR leads to less sparse, thus less binarizable representation at the backbone level. }
    \label{fig:simclr_temperature}
\end{figure}

\paragraph{The importance of hyper-parameters} There is a temperature hyper-parameter in SimCLR that control the sharpness of the softmax in the loss. We investigate in \cref{fig:simclr_temperature}, the impact of this hyper-parameter on the sparsity and the binarizarion easiness of the backbone representation. We show that when using high temperatures with SimCLR, the backbone representation get less binarizable.

\section{Proof: SimCLR representations are binarizable when using a large backbone dimension}
\label{sec:proof}

If one has roughly symmetrical, centered and independent distributions between the embeddings dimensions, then performing quantization will only collapse different images to the same quantized code with very low probability --going exponentially quickly to $0$ with the embedding dimension. We demonstrate this as follows:
\begin{align*}
    p(&\|Q(z_i)-Q(z_j)\|=0)=p(\bigwedge_{k=1}^{K}Q(z_i)_k=Q(z_j)_k)\\
    =&\prod_{k=1}^{K}p(Q(z_i)_k=Q(z_j)_k)\\
    =&\prod_{k=1}^{K}p(z_{i,k}>0 \wedge z_{j,k}>0 \cup z_{i,k}<0 \wedge z_{j,k}<0)\\
    =&\prod_{k=1}^{K}p(z_{i,k}>0 \wedge z_{j,k}>0)+p(z_{i,k}<0 \wedge z_{j,k}<0)\\
    =&0.5^{K},
\end{align*}
Because of the above, it is clear that as $K$ increases as none of the (random) embeddings will be matched to the same hypercube vertex i.e. it is possible to learn an inverse mapping, and all the information about the embeddings is kept after quantization.
The second step of our argument allows us to understand why does the above holds even for trained embeddings e.g. using SimCLR. In fact, in practice all the embeddings are not randomly distributed and in fact if training is successful, we have that in an ideal scenario $\|z_i-z_j\|\approx 0$ if $x_i=T(x_j)$. Nevertheless, it was shown in \cite{wang2020hypersphere} that modulo those augmentations, for which SimCLR learns to be invariant to, the embeddings are uniformly distributed on the sphere. That is, quantization will naturally preserve the already learned invariance, and preserve all the information that the DN's was not invariant too and for which the information is random on the sphere. In consequence,
\begin{itemize}
    \item If the representations are as uniform as possible on the sphere, then quantization will have minimal impact on performances (especially in high dimension cf \cref{sec:good}).
    \item If you are given two DNs with same output dimension, the one that at the least uniform embeddings is the one whose information content will be further removed through the quantization step
\end{itemize}


\begin{table}[ht]
    \footnotesize
    \centering
    \begin{tabular}{|r|c|c|c|c|c|c|c|c|c|c|c|}
        \hline
         \textbf{Dataset} & \textbf{Model} & \multicolumn{10}{c|}{\textbf{Backbone Dimension}}\\
         Dataset & Model & 512 & 1024 & 2048 & 4096 & 8192 & 10240 & 12288 & 15360 & 20480 & 32768\\
        \hline
         Imagenet & SimCLR & 61.85 & 64.41 & 66.28 & 68.11 & 69.22 & 69.68 & 69.9 & 70.09 & 70.51 & 71.08\\
          & VICReg & 63.81 & 66.35 & 68.34 & 70.01 & 71.22 & 71.37 & 71.75 & 71.84 & 71.95 & 72.21 \\
          & Byol & 57.82 & 60.68 & 62.89 & 64.76 & 65.59 & 66.38 & 66.68 & 67.47 & 68.4 & 68.07\\
          & Supervised & \textbf{76.24} & 76.0 & 75.47 & 75.23 & 75.03 & 75.07 & 74.91 & 74.82 & 74.78 & 74.69 \\
         \hline
         Imagenet 10\% & SimCLR & 51.36 & 53.13 & 54.31 & 55.55 & 56.51 & 56.71 & 56.75 & 57.18 & 57.39 & \textbf{58.33}\\
          & VICReg & 54.72 & 56.7 & 57.96 & 59.13 & 59.88 & 60.08 & \textbf{60.49} & 60.38 & 60.3 & 60.45 \\
          & Byol & 43.11 & 44.81 & 45.94 & 47.71 & 47.95 & 49.35 & 49.57 & 50.61 & \textbf{52.9} & 52.37 \\
          & Supervised & \textbf{73.12} & 72.32 & 71.36 & 70.13 & 69.27 & 68.93 & 68.69 & 68.58 & 68.27 & 67.99 \\
        \hline
         Imagenet 1\% & SimCLR & 28.85 & 30.21 & 31.41 & 32.19 & 33.0 & 33.12 & 33.46 & 33.75 & 34.25 & 35.3 \\
          & VICReg & 33.33 & 34.46 & 34.98 & 35.92 & 36.51 & 36.44 & 36.67 & 36.8 & 36.67 & 37.05 \\
          & Byol & 21.63  & 23.74  & 25.18  & 27.2  & 28.99  & 29.46  & 30.08  & 30.9  & 32.24  & 33.58 \\
          & Supervised & \textbf{56.38} & 55.36 & 54.06 & 51.98 & 50.24 & 49.51 & 49.25 & 48.86 & 48.27 & 47.88 \\
        \hline
        CIFAR10 & SimCLR & 77.04 & 79.26 & 80.92 & 82.45 & 82.66 & 83.42 & 83.75 & 83.54 & 84.01 & \textbf{85.08} \\
        & VICReg & 76.76 & 79.84 & 81.09 & 83.2 & 85.25 & 85.1 & 86.1 & 85.77 & 86.07 & \textbf{86.7} \\ 
        & Byol & 68.76 & 71.13 & 72.62 & 75.21 & 75.32 & 76.81 & 77.05 & 78.41 & \textbf{78.93} & 78.71 \\
        & Supervised & 77.19 & 80.42 & 82.84 & 84.96 & 86.83 & 87.41 & 87.7 & 88.63 & 88.74 & \textbf{89.38} \\
        \hline
        CIFAR100 & SimCLR & 51.86 & 54.75 & 57.07 & 59.61 & 59.7 & 60.81 & 61.27 & 61.32 & 61.83 & \textbf{63.21} \\
        & VICReg & 51.93 & 54.77 & 58.47 & 61.01 & 63.94 & 63.69 & 64.69 & 65.47 & 65.12 & 66.9 \\ 
        & Byol & 41.19 & 44.31 & 47.1 & 49.96 & 50.34 & 52.1 & 52.72 & 54.0 & 54.65 & 54.68 \\
        & Supervised & 52.94 & 58.06 & 62.03 & 64.49 & 67.93 & 69.11 & 69.88 & 71.02 & 71.42 & 72.62 \\
        \hline
        CLEVR-D & SimCLR & 43.43 & 44.74 & 46.85 & 48.06 & 50.13 & 50.59 & 49.91 & 51.59 & 51.73 & \textbf{53.72} \\
        & VICReg & 42.12 & 41.97 & 44.75 & 46.96 & 48.03 & 49.74 & 49.73 & 50.92 & 51.09 & \textbf{52.75} \\
        & Byol & 36.41 & 38.27 & 38.56 & 40.35 & 40.47 & 42.0 & 42.91 & 45.44 & 45.94 & \textbf{47.93}\\
        & Supervised & 41.74 & 44.63 & 46.37 & 50.21 & 51.97 & 52.33 & 52.31 & 53.15 & 53.11 & \textbf{54.38} \\
        \hline
        CLEVR-C & SimCLR & 44.78 & 46.73 & 47.84 & 48.88 & 49.98 & 50.94 & 51.03 & 51.13 & 49.86 & 51.09 \\
        & VICReg & 42.35 & 44.28 & 46.77 & 48.77 & 49.17 & 50.81 & 50.77 & 51.95 & 52.17 & 52.07 \\
        & Byol & 35.73 & 37.92 & 38.31 & 38.42 & 38.73 & 39.95 & 42.22 & 44.27 & 44.17 & 44.51\\
        & Supervised & 40.09 & 45.79 & 48.37 & 51.87 & 51.88 & 53.43 & 53.31 & 54.76 & 55.0 & 55.32\\
        \hline
        Places & SimCLR & 46.62 & 49.0 & 50.77 & 52.59 & 53.61 & 54.04 & 54.17 & 54.63 & 54.83 & \textbf{55.54} \\
        & VICReg & 46.8 & 49.0 & 50.97 & 52.74  & 54.22 & 54.48 & 54.9 & 55.17 & 55.49 & \textbf{55.96} \\
        & Byol & 44.13 & 46.37 & 48.36 & 50.18 & 50.72 & 51.44 & 51.94 & 52.57 & 53.34 & \textbf{53.39} \\
        & Supervised & 47.98 & 50.03 & 51.9 & 53.53 & 54.3 & 54.46 & 55.03 & 55.09 & 55.49 & \textbf{55.65} \\
        \hline
        Eurosat & SimCLR & 87.26 & 88.58 & 88.52 & 88.2 & 89.96 & 89.7 & 89.8 & 90.48 & 90.16 & \textbf{91.68} \\
        & VICReg & 86.14 & 89.1 & 89.96 & 90.88 & 92.04 & 92.34 & 91.68 & 93.04 & 92.52 & \textbf{93.16} \\ 
        & Byol & 78.74 & 83.0 & 84.04 & 85.48 & 86.3 & 86.3 & 86.52 & 88.1 & 87.8 & \textbf{88.28} \\
        & Supervised & 84.46 & 87.2 & 88.82 & 91.54 & 93.64 & 94.44 & 94.42 & 94.96 & 95.32 & \textbf{95.92} \\
        \hline 
    \end{tabular}
    \caption{Linear probe accuracy for SimCLR, VicReg, Byol and a supervised model across several downstream tasks (Imagenet-1k using only 10\% of the training examples, CIFAR10, CLEVR-Distance, Places205 and eurosat). Even if the supervised model get worse performances on ImageNet when increasing the size of the backbone, it's worth to note that when looking at other downstream tasks the performances increase significantly when using wider backbone. This showed that even in the supervised learning case, increasing the dimension of the backbone increase significantly the robustness with respect to the pretraining bias. }
    \label{tab:downstream_tasks_all}
\end{table}

\end{document}